\documentclass[runningheads]{llncs}
\usepackage{graphicx}
\usepackage{amsmath,amssymb} 
\usepackage{color}

\usepackage{subcaption}
\usepackage{overpic}

\usepackage[ruled]{algorithm2e}
\usepackage{algorithmic}

\usepackage{tabularx}
\usepackage{booktabs}

\begin{document}
\pagestyle{headings}
\mainmatter

\def\ACCV20SubNumber{286}  

\title{DEAL: Difficulty-aware Active Learning for Semantic Segmentation} 
\titlerunning{Difficulty-aware Active Learning for Semantic Segmentation}
%
\author{
Shuai Xie\inst{1} \and
Zunlei Feng\inst{1} \and
Ying Chen\inst{1} \and
Songtao Sun\inst{1} \and
Chao Ma\inst{1} \and
Mingli Song\inst{1}
}
\authorrunning{S. Xie et al.}
%
\institute{
Zhejiang University, Hangzhou, China \\
\email{\{shuaixie, zunleifeng, lynesychen, songtaosun, chaoma, brooksong\}@zju.edu.cn}
}
\maketitle

\begin{abstract}
Active learning aims to address the paucity of labeled data by finding the most informative samples. However, when applying to semantic segmentation, existing methods ignore the segmentation difficulty of different semantic areas, which leads to poor performance on those hard semantic areas such as tiny or slender objects. To deal with this problem, we propose a semantic Difficulty-awarE Active Learning (DEAL) network composed of two branches: the common segmentation branch and the semantic difficulty branch. For the latter branch, with the supervision of segmentation error between the segmentation result and GT, a pixel-wise probability attention module is introduced to learn the semantic difficulty scores for different semantic areas. Finally, two acquisition functions are devised to select the most valuable samples with semantic difficulty. Competitive results on semantic segmentation benchmarks demonstrate that DEAL achieves state-of-the-art active learning performance and improves the performance of the hard semantic areas in particular.
\end{abstract}


\section{Introduction}
Semantic segmentation is a fundamental task for various applications such as autonomous driving \cite{teichmann2018multinet,feng2020deep}, biomedical image analysis \cite{ronneberger2015u,yang2017suggestive,zheng2019new}, remote sensing \cite{azimi2019skyscapes} and robot manipulation \cite{puang2019visual}.
Recently, data-driven methods have achieved great success with large-scale datasets \cite{cordts2016cityscapes,zhou2017scene}. However, tremendous annotation cost has become an obstacle for these methods to be widely applied in practical scenarios.
Active Learning (AL) can be the right solution by finding the most informative samples. Annotating those selected samples can support sufficient supervision information and reduce the requirement of labeled samples dramatically.

Previous methods can be mainly categorized into two families: uncertainty-based \cite{wang2016cost,gorriz2017cost,gal2017deep,yoo2019learning} and representation-based \cite{sener2017active,sinha2019variational,gissin2019discriminative}.
However, many works \cite{wang2016cost,gal2017deep,sener2017active,gissin2019discriminative} are only evaluated on image classification benchmarks. 
There has been considerably less work specially designed for semantic segmentation.
Traditional uncertainty-based methods like Entropy \cite{shannon1948mathematical} and Query-By-Committee (QBC) \cite{seung1992query} have demonstrated their effectiveness in semantic segmentation \cite{kuo2018cost,Casanova2020Reinforced}.
However, all of them are solely based on the uncertainty reflected on each pixel, without considering the semantic difficulty and the actual labeling scenarios.

\begin{figure}[t]
    \centering
    \begin{subfigure}{0.33\columnwidth}
        \centering
        \includegraphics[width=\linewidth]{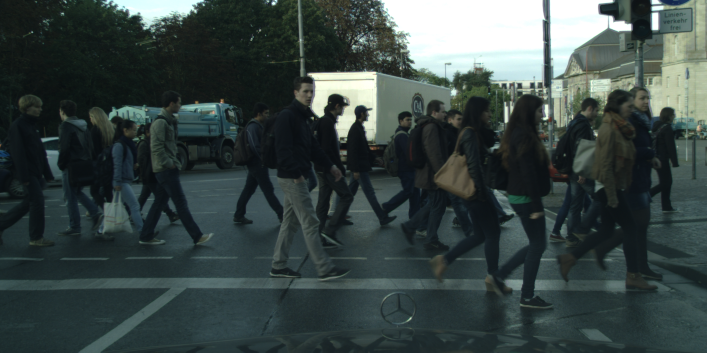}
        \caption{Image}
    \end{subfigure}
    \begin{subfigure}{0.33\columnwidth}
        \centering
        \includegraphics[width=\linewidth]{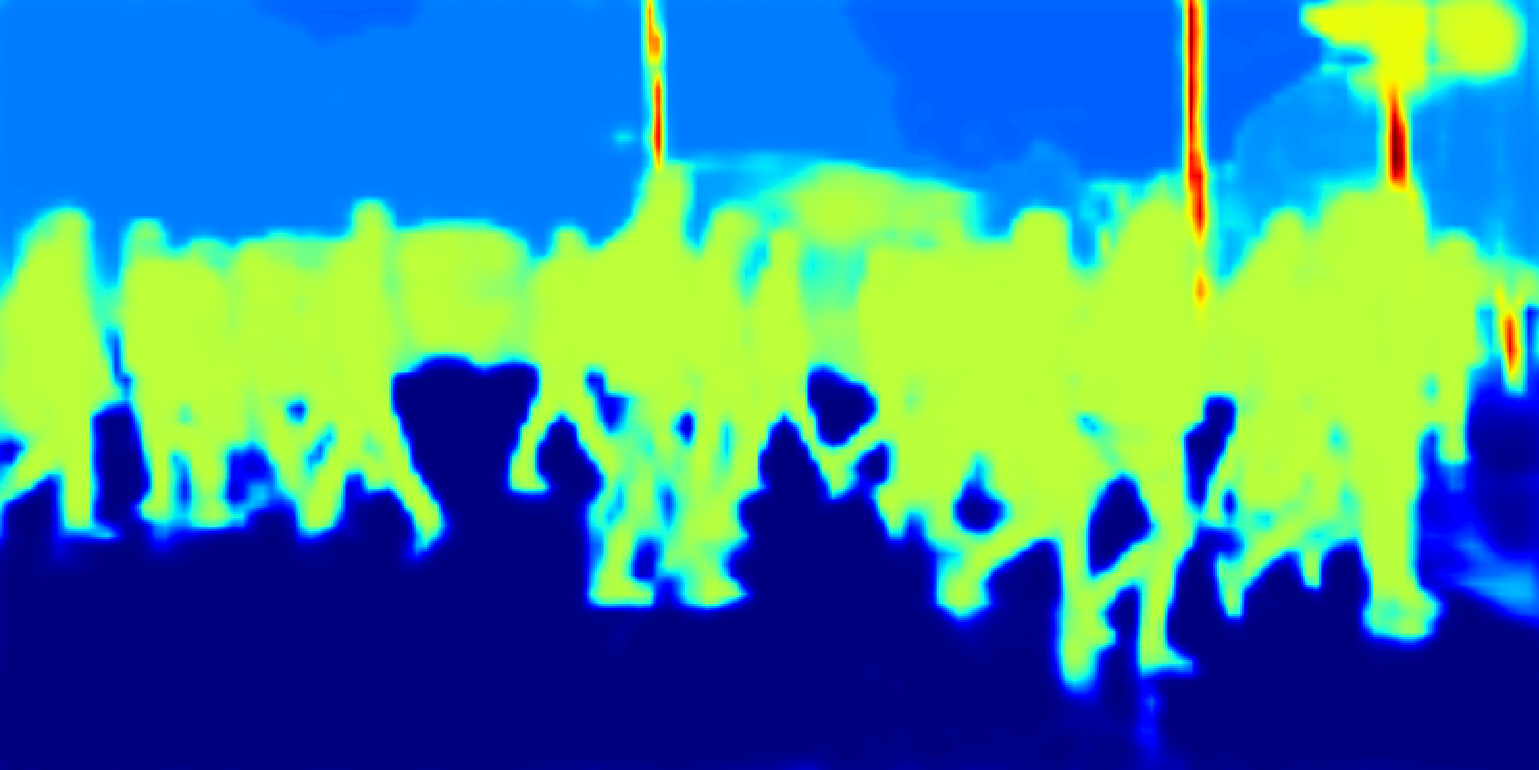}
        \caption{Semantic difficulty map}
    \end{subfigure}

    \begin{subfigure}{0.3\columnwidth}
        \centering
        \includegraphics[width=\linewidth]{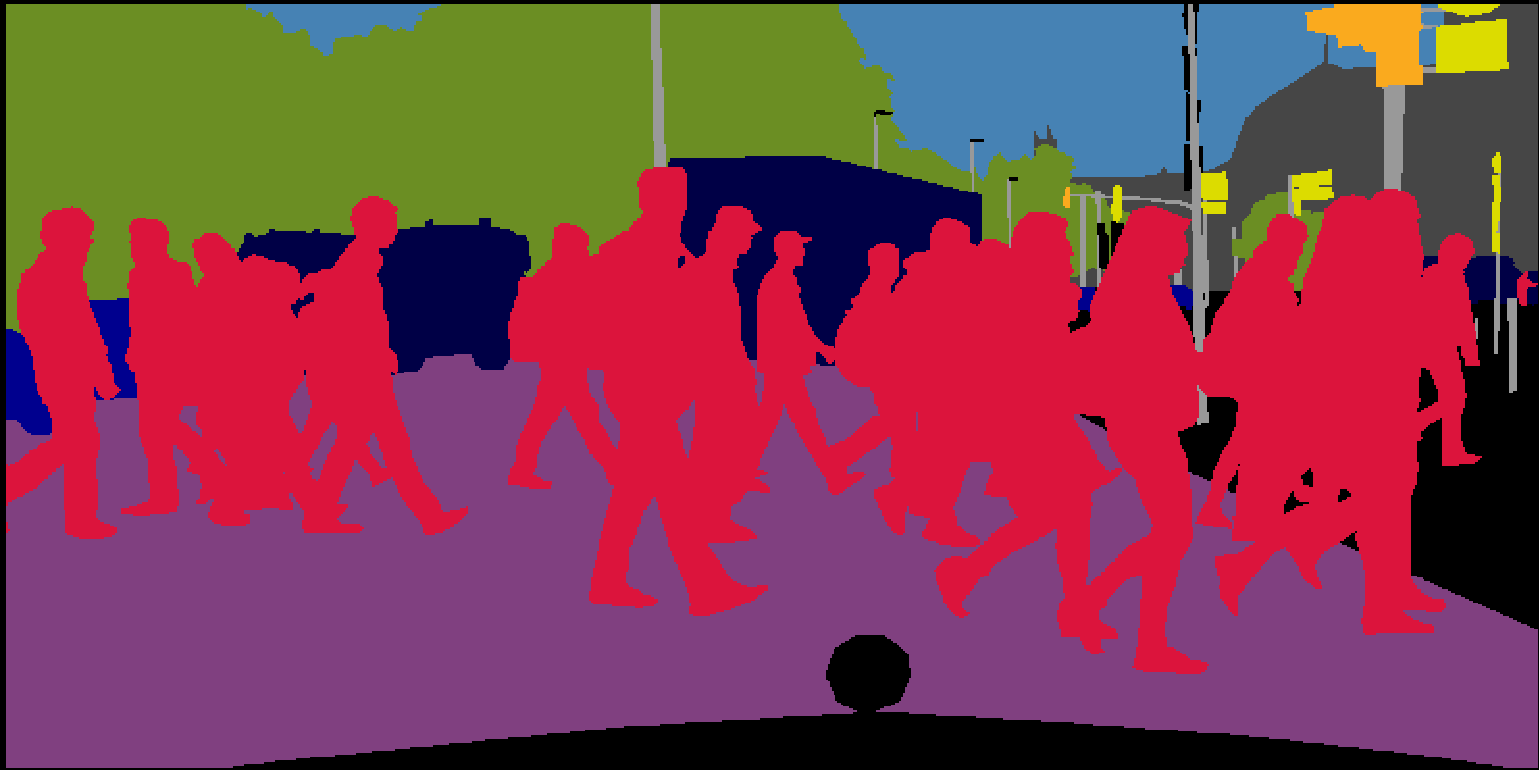}
        \caption{GT}
    \end{subfigure}
        \begin{subfigure}{0.3\columnwidth}
        \centering
        \includegraphics[width=\linewidth]{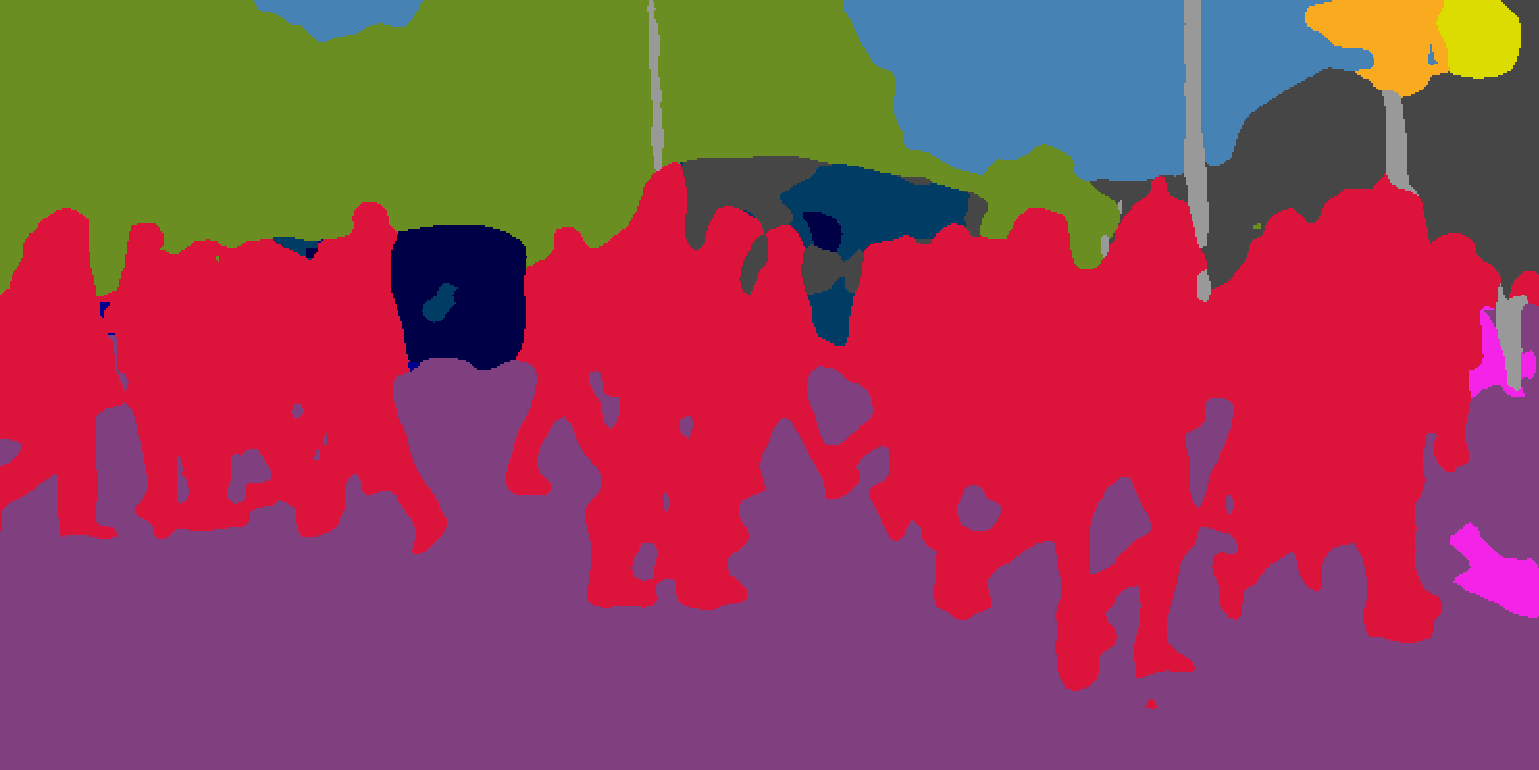}
        \caption{Predicted result}
    \end{subfigure}
    \begin{subfigure}{0.3\columnwidth}
        \centering
        \includegraphics[width=\linewidth]{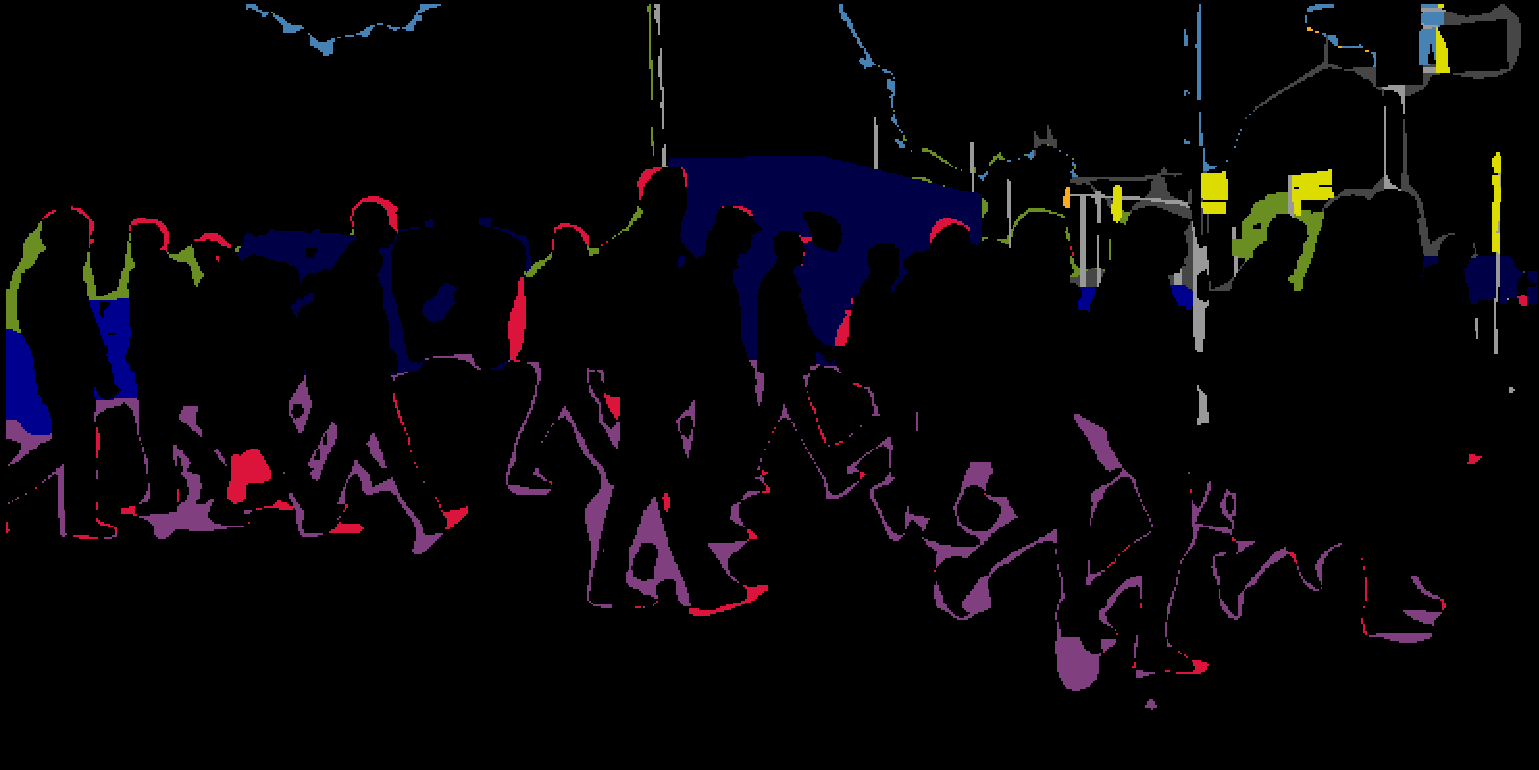}
        \caption{Error mask}
    \end{subfigure}
    \caption{
        (a) Image from Cityscapes \cite{cordts2016cityscapes}.
        (b) Semantic difficulty map. The colder color represents the easier semantic areas such as road, sky, and buildings. The warmer color represents the harder semantic areas such as poles and traffic signs.
        (c) GT.
        (d) Predicted result.
        (e) Error mask generated with (c) and (d)
        according to Eq. (\ref{equ:mask}). It's a binary image, coloring for better visualization.
    }
    \label{fig:intro}
\end{figure}

In this paper, we propose a semantic Difficulty-awarE Active Learning (DEAL) method taking the semantic difficulty into consideration.
Due to the class imbalance and shape disparity, a noticeable semantic difficulty difference exists among the different semantic areas in an image. To capture this difference, we adopt a two-branch network composed of a semantic segmentation branch and a semantic difficulty branch.
For the former branch, we adopt the common segmentation network.
For the latter branch, we leverage the wrong predicted result as the supervision, which is termed as the \textit{error mask}. It's a binary image where the right and wrong positions have a value 0 and 1, respectively. As illustrated in Fig. \ref{fig:intro}(e), we color these wrong positions for better visualization. 
Then, a pixel-wise probability attention module is introduced to aggregate similar pixels into areas and learn the proportion of misclassified pixels as the difficulty score for each area. Finally, we can obtain the semantic difficulty map in Fig. \ref{fig:intro}(b).

Then two acquisition functions are devised based on the map.
One is Difficulty-aware uncertainty Score (DS) combining the uncertainty and difficulty.
The other is Difficulty-aware semantic Entropy (DE) solely based on the difficulty.
Experiments show that 
the learned difficulty scores have a strong connection with the standard evaluation metric IoU.
And DEAL can effectively improve the overall AL performance and the IoU of the hard semantic classes in particular.

In summary, our major contributions are as follows:
1) Proposing a new AL framework incorporating the semantic difficulty to select the most informative samples for semantic segmentation.
2) Utilizing \textit{error mask} to learn the semantic difficulty.
3) Competitive results on CamVid \cite{brostow2009semantic} and Cityscapes \cite{cordts2016cityscapes}.


\section{Related Work}

\subsubsection{AL for semantic segmentation}

The core of AL is measuring the informativeness of the unlabelled samples. 
Modern AL methods can be mainly divided into two groups: uncertainty-based  \cite{wang2016cost,gorriz2017cost,gal2017deep,yoo2019learning} and representation-based \cite{sener2017active,sinha2019variational,gissin2019discriminative}. The latter views the AL process as an approximation of the entire data distribution and query samples to increase the data diversity, such as Core-set \cite{sener2017active} and VAAL \cite{sinha2019variational}, which can be directly used in semantic segmentation.
There are also some methods specially designed for semantic segmentation, which can also be divided into two groups: image-level \cite{yang2017suggestive,gorriz2017cost,kuo2018cost,tan2019batch} and region-level \cite{mackowiak2018cereals,siddiqui2020viewal,Casanova2020Reinforced}.

Image-level methods use the complete image as the sampling unit. \cite{yang2017suggestive} propose suggestive annotation (SA) and train a group of models on various labeled sets 
obtained with bootstrap sampling 
and select samples with the highest variance.
\cite{gorriz2017cost} employ MC dropout to measure uncertainty for melanoma segmentation.
\cite{kuo2018cost} adopt QBC strategy and propose a cost-sensitive active learning method for intracranial hemorrhage detection.
\cite{tan2019batch} build a batch mode multi-clue method, incorporating edge information with QBC strategy and graph-based representativeness.
All of them are based on a group of models and time-consuming when querying a large unlabeled data pool.

Region-level methods only sample the informative regions from images.
\cite{mackowiak2018cereals} combines the MC dropout uncertainty with an effort estimation regressed from the annotation click patterns, which is hard to access for many datasets.
\cite{siddiqui2020viewal} propose ViewAL and use the inconsistencies in model predictions across viewpoints to measure the uncertainty of super-pixels, which is specially designed for RGB-D data.
\cite{Casanova2020Reinforced} model a deep Q-network-based query network as a reinforcement learning agent, trying to learn sampling strategies based on prior AL experience. 
In this work, we incorporate the semantic difficulty to measure the informativeness and select samples at the image level. Region-level method will be studied in the future.

\subsubsection{Self-attention mechanism for semantic segmentation}
The self-attention mechanism is first proposed by \cite{vaswani2017attention} in the machine translation task.
Now, it has been widely used in many tasks \cite{vaswani2017attention,Lin2017ASS,wang2016attention,zhou2018atrank} owing to its intuition, versatility and interpretability \cite{Chaudhari2019AnAS}.
The ability to capture the long-range dependencies inspires many semantic segmentation works designing their attention modules. 
\cite{zhao2018psanet} use a point-wise spatial attention module to aggregate context information in a self-adaptive manner.
\cite{yuan2018ocnet} introduce an object context pooling scheme to better aggregate similar pixels belonging to the same object category.
\cite{Huang2019CCNetCA} replace the non-local operation \cite{Wang2018NonlocalNN} into two consecutive criss-cross operations and gather long-range contextual
information in the horizontal and vertical directions. 
\cite{fu2019dual} design two types of attention modules to exploit the dependencies between pixels and channel maps.
Our method also uses the pixel-wise positional attention mechanism in \cite{fu2019dual} to aggregate similar pixels.


\section{Method}

Before introducing our method, we first give the definition of the AL problem.
Let $(x^a, y^a)$ be an annotated sample from the initial annotated dataset $\mathcal{D}^a$ and $x^u$ be an unlabeled sample from a much larger unlabeled data pool $\mathcal{D}^u$.
AL aims to iteratively query a subset $\mathcal{D}^s$ containing the most informative $m$ samples $\{x^u_1,x^u_2,..., x^u_m\}$ from $\mathcal{D}^u$, where $m$ is a fixed budget.

In what follows, we first give an overview of our difficulty-aware active learning framework, then detail the probability attention module and loss functions, finally define two acquisition functions.

\subsection{Difficulty-aware Active Learning}
\begin{figure}[t]
    \centering
    \begin{overpic}[width=1.\linewidth]{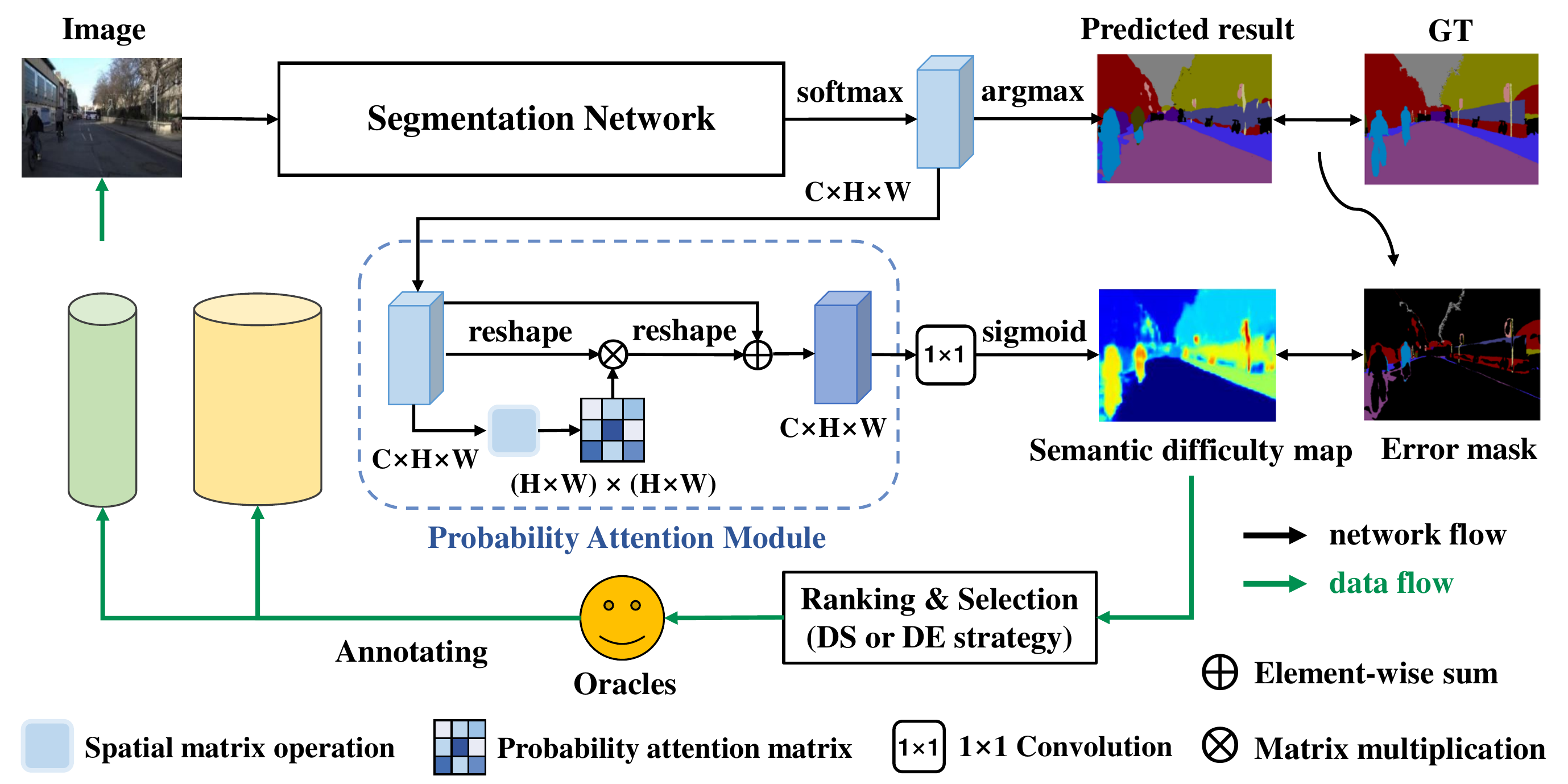}
    \put(2.2,32.2){\fontsize{8}{8}$(x^a, y^a)$}
    \put(15,32.2){\fontsize{8}{8}$x^u$}
    \put(5,25){\fontsize{8}{8}$\mathcal{D}^a$}
    \put(15,25){\fontsize{8}{8}$\mathcal{D}^u$}
    \put(11,8){\fontsize{8}{8}$\mathcal{D}^s$}
    
    \put(8,13.5){\fontsize{10}{10}$+$}
    \put(18.2,13.5){\fontsize{10}{10}$-$}
    
    \put(61,37){\fontsize{8}{8}$S^{*}$}
    \put(75.,36){\fontsize{8}{8}$S^{p}$}
    \put(91.2,36){\fontsize{8}{8}$S^{g}$}
    \put(75,32.5){\fontsize{8}{8}$M^{d}$}
    \put(91,32.5){\fontsize{8}{8}$M^{e}$}
    
    \put(27,32.4){\fontsize{8}{8}$P$}
    \put(53,32.4){\fontsize{8}{8}$Q$}
    
    \put(81.5,43.7){\fontsize{8}{8}$\mathcal{L}_{seg}$}
    \put(81.7,29){\fontsize{8}{8}$\mathcal{L}_{dif}$}
    \end{overpic}
    \caption{
    Overview of our difficulty-aware active learning framework for semantic segmentation.
    The first branch is a common semantic segmentation network.
    The second branch is composed of a praobability attention module and a $1 \times 1$ convolution.
    $\mathcal{D}^{a}$ and $\mathcal{D}^{u}$ are the annotated and unlabeled data, $\mathcal{D}^{s}$ is a subset selected from $\mathcal{D}^{u}$.
    $P$ and $Q$ are the probability maps before and after attention.
    $\mathcal{L}_{seg}$ and $\mathcal{L}_{dif}$ are the loss functions described in Eq. \ref{equ:seg} and Eq. \ref{equ:dif}.
    DS and DE strategies are detailed in Sec. \ref{sec:acquisition}.
    }
    \label{fig:overview}
\end{figure}
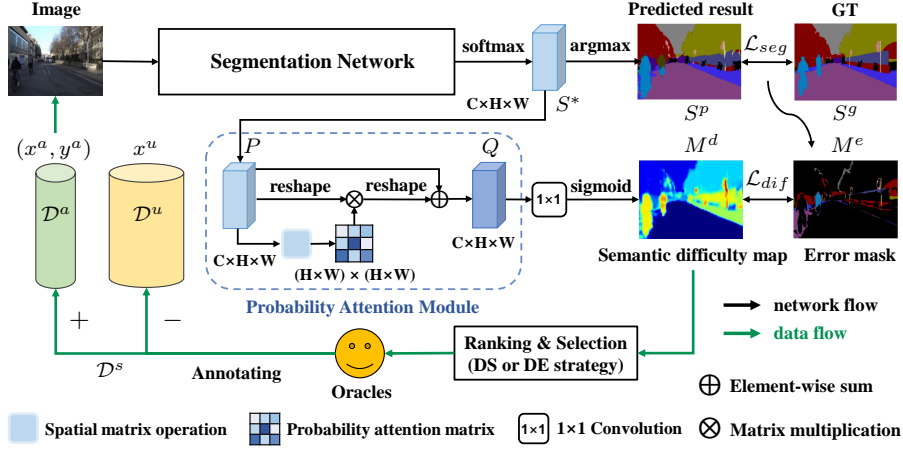

To learn the semantic difficulty, we exploit the \textit{error mask} generated from the segmentation result.
Our intuition is that these wrong predictions are what our model ``feels'' difficult to segment, which may have a relation with the semantic difficulty.
Thus, we build a two-branch network generating semantic segmentation result and semantic difficulty map in a multi-task manner, best viewed in Fig. \ref{fig:overview}. 
The first branch is a common segmentation network, which can be used to generate the \textit{error mask}. The second branch is devised to learn the semantic difficulty map with the guidance of \textit{error mask}.
This two-branch architecture is inspired by \cite{yoo2019learning}. In their work, a loss prediction module is attached to the task model to predict a reliable loss for $x^u$ and samples with the highest losses are selected. While in our task, we dig deeper into the scene and analyze the semantic difficulty of each area.

As illustrated in Fig. \ref{fig:overview}, the first branch can be any semantic segmentation network. Assume $S^*$ is the output of softmax layer and $S^p$ is the prediction result after the argmax operation. With the segmentation result $S^{p}$ and GT $S^{g}$, the \textit{error mask} $M^{e}$ can be computed by:
\begin{align} \label{equ:mask} 
    M^{e}_k = \begin{cases}
    1 & \text{if } S^{p}_k \neq S^{g}_k,\\
    0 & \text{otherwise},
    \end{cases}
\end{align}
where $S^{p}_k$ and $S^{g}_k$ denote the $k^{th}$ pixel value of the segmentation result and GT, $M^{e}_k$ is the $k^{th}$ pixel value of the \textit{error mask}.

The second branch is composed of two parts: a probability attention module and a simple $1 \times 1$ convolution layer used for binary classification.
The softmax output of the first branch $S^*$ is directly used as the input of the second branch, which are $C$-channel probability maps and $C$ is the number of classes. We denote it as $P \in \mathcal{R}^{C \times H \times W}$ and $P_k$ is the probability vector for the $k^{th}$ pixel.
Using probability maps is naive but accompanied with two advantages.
First, pixels with similar difficulty tend to have similar $P_k$.
Second, pixels of the same semantic tend to have similar $P_k$.
Combined with a pixel-wise attention module, we can easily aggregate these similar pixels and learn similar difficulty scores for them.
In our learning schema, the performance of the second branch depends much on the output of the first branch. However, there is no much difference if we branch these two tasks earlier and learn the independent features.
We validate this opinion in Sec. \ref{sec:branch_pos}.

The semantic difficulty learning process can be imagined into two steps.
Firstly, we learn a binary segmentation network with the supervision of \emph{error mask} $M^{e}$. Each pixel will learn a semantic difficulty score.
Secondly, similar pixels are aggregated into an area so that this score can be spread among them. Finally, we can obtain the semantic difficulty map $M^d$.

\subsection{Probability Attention Module}
In this section, we detail the probability attention module (PAM) in our task.
Inspired by \cite{fu2019dual}, we use this module to aggregate pixels with similar softmax probability.
Given the probability maps $P \in \mathcal{R}^{C \times H \times W}$, we first reshape it to $P \in \mathcal{R}^{C \times K}$, where $K=H \times W$.
Then the probability attention matrix $A \in \mathcal{R}^{K \times K}$ can be computed with $P^TP$ and a softmax operation as below:
\begin{align} \label{equ:att}
\begin{split}
A_{ji} &= \frac{exp(P^T_i \cdot P_j)}{\sum_{i=1}^K exp(P^T_i \cdot P_j)}, \\
Q_j &= \gamma \sum_{i=1}^K (A_{ji}P_i) + P_j,
\end{split}
\end{align}
where $A_{ji}$ is the $i^{th}$ pixel's impact on $j^{th}$ pixel, $P_j$ is the original probability vector of the $j^{th}$ pixel and $Q_j$ is the one after attention, $\gamma$ is a learnable weight factor.
Finally, we can get the probability maps $Q \in \mathcal{R}^{C \times H \times W}$ after attention.

\begin{figure}[t]
    \centering
    \begin{subfigure}{0.4\textwidth}
        \centering
        \includegraphics[width=0.5\textwidth]{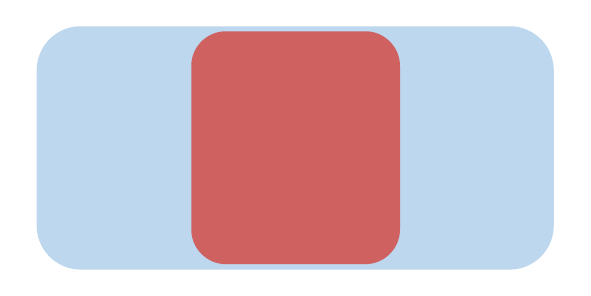}
        \caption{Error inside the object}
        \label{fig:error_a}
    \end{subfigure}
    \begin{subfigure}{0.4\textwidth}
        \centering
        \includegraphics[width=0.5\textwidth]{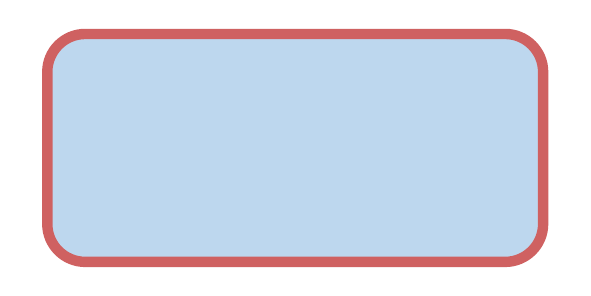}
        \caption{Error on the object boundary}
        \label{fig:error_b}
    \end{subfigure}
    \caption{Two typical errors in semantic segmentation. The right and wrong areas are in blue and red, best viewed in color.}
    \label{fig:error}
\end{figure}

\begin{figure}[b]
    \centering
    \includegraphics[width=1.0\textwidth]{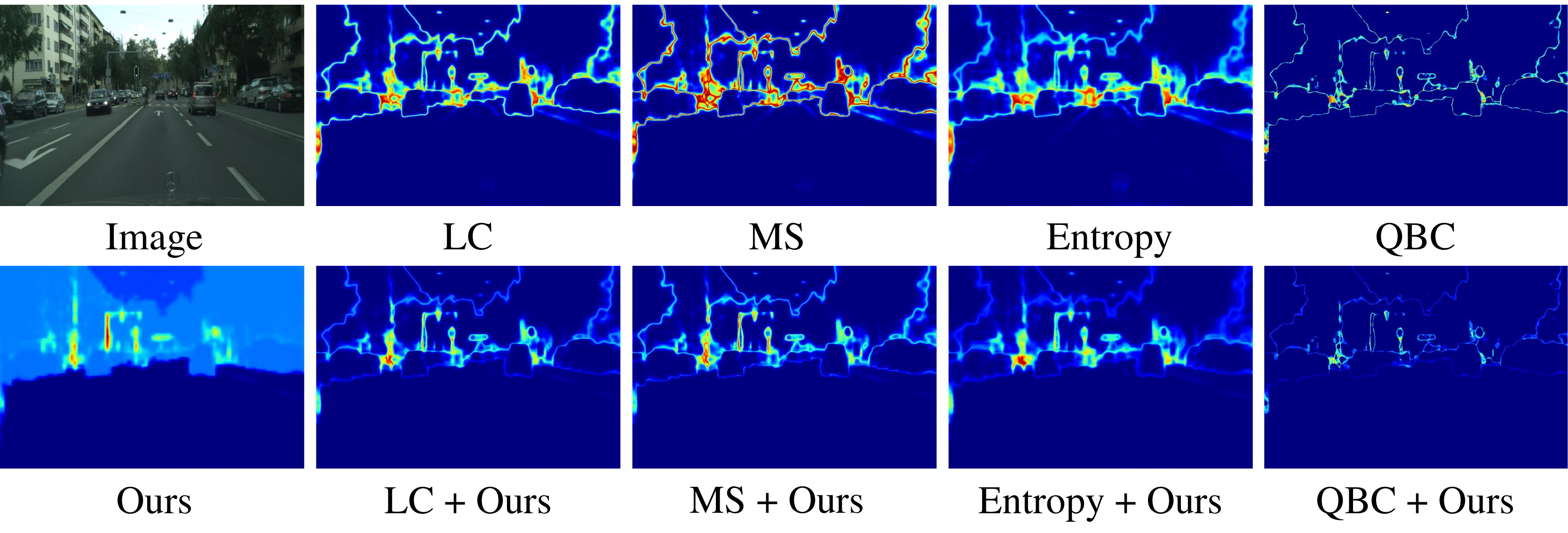}
    \caption{First row: Image from Cityscapes \cite{cordts2016cityscapes} and traditional uncertainty maps. Second row: Our semantic difficulty map and difficulty-aware uncertainty maps. The warmer color means the higher uncertainty.}
    \label{fig:uncer_cmp}
\end{figure}

Let's take the segmentation result of the two bicyclists in Fig. \ref{fig:overview} to explain the role of PAM as it reflects two typical errors in semantic segmentation: (1) error inside the object (the smaller one $b_1$); (2) error on the object boundary (the larger one $b_2$), as shown in Fig. \ref{fig:error}.
Assume our attention module can aggregate pixels from the same object together, the right part of the object learns 0 while the wrong part learns 1.
Since $b_1$ has a larger part of wrong areas, it tends to learn larger difficulty scores than $b_2$.
Similar to objects, pixels from the same semantic, such as road, sky and buildings can also learn similar difficulty scores.
Ablation study in Sec. \ref{sec:pam} also demonstrates that PAM can learn more smooth difficulty scores for various semantic areas.

Some traditional methods also employ the softmax probabilities to measure the uncertainty, such as least confidence (LC) \cite{settles2009active}, margin sampling (MS) \cite{scheffer2001active} and Entropy \cite{shannon1948mathematical}.
The most significant difference between our method and these methods is that we consider difficulty at the semantic level with an attention module, rather than measuring the uncertainty of each pixel alone. 
QBC \cite{seung1992query} can use a group of models, but it still stays at the pixel level.
To clearly see the difference, we compare our semantic difficulty map with the uncertainty maps of these methods in Fig. \ref{fig:uncer_cmp}. 
The first row are uncertainty maps generated with these methods, which are loyal to the uncertainty of each pixel.
For example, some pixels belonging to \emph{sky} can have the same uncertainty with \emph{traffic light} and \emph{traffic sign}.
Supposing an image has many pixels with high uncertainty belonging to the easier classes, it will be selected by these methods.
While our semantic difficulty map (first in the second row) can serve as the difficulty attention and distinguish more valuable pixels.
As illustrated in the second row, combined with our difficulty map, the uncertainty of the easier semantic areas like \emph{sky} is suppressed while the harder semantic areas like \emph{traffic sign} is reserved.

\subsection{Loss Functions}

\subsubsection{Loss of Semantic Segmentation}
To make an equitable comparison with other methods, we use the standard cross-entropy loss for the semantic segmentation branch, which is defined as:
\begin{align} \label{equ:seg}
    \mathcal{L}_{seg}(S^*, S^{g}) = \frac{1}{K} \sum^{K}_{k=1} \mathcal{\ell}(S_k^*, S_k^{g}) + R(\theta),
\end{align}
where $S_k^*$ and $S_k^{g}$ are the segmentation output and ground truth for pixel $k$, $\mathcal{\ell}(\cdot)$ is the cross-entropy loss, $K$ is the total pixel number, and $R$ is the L2-norm regularization term.

\subsubsection{Loss of Semantic Difficulty}
For the semantic difficulty branch, we use an inverted weighted binary cross-entropy loss defined below, as there is a considerable imbalance between the right and wrong areas of \emph{error mask}.
\begin{align} \label{equ:dif}
\begin{split}
    \mathcal{L}_{dif}(M^d, M^e) = -&\frac{1}{K} \sum_{k=1}^K \lambda_1 M_k^e log(M_k^d) + \lambda_2 (1-M_k^e)log(1-M_k^d), \\
    &\lambda_1 = \frac{\sum_k^K \mathbf{1}(M_k^e=0)}{K}, \lambda_2 = 1 - \lambda_1,
\end{split}
\end{align}
where $M_k^d$ and $M_k^e$ are the difficulty prediction and \emph{error mask} ground truth for pixel $k$, $\mathbf{1(\cdot)}$ is the indicator function, and $\lambda_1$ and $\lambda_2$ are dynamic weight factors.

\subsubsection{Final Loss}
Our final training objective is a combination of Eq. \ref{equ:seg} and Eq. \ref{equ:dif}, which is computed as:
\begin{align} \label{equ:sum}
    \mathcal{L} = \mathcal{L}_{seg} + \alpha \mathcal{L}_{dif},
\end{align}
where $\alpha$ is a weight factor and set to 1 in the experiments.

\subsection{Acquisition Functions}
\label{sec:acquisition}

Samples from $\mathcal{D}^u$ are usually ranked with a scalar score in AL.
However, semantic segmentation is a dense-classification task, many methods output a score for each pixel on the image, including our semantic difficulty map.
Thus, it's quite significant to design a proper acquisition function.
Below are two functions we have designed.

\begin{figure}[t]
    \centering
    \begin{subfigure}{0.325\textwidth}
        \centering
        \includegraphics[width=\textwidth]{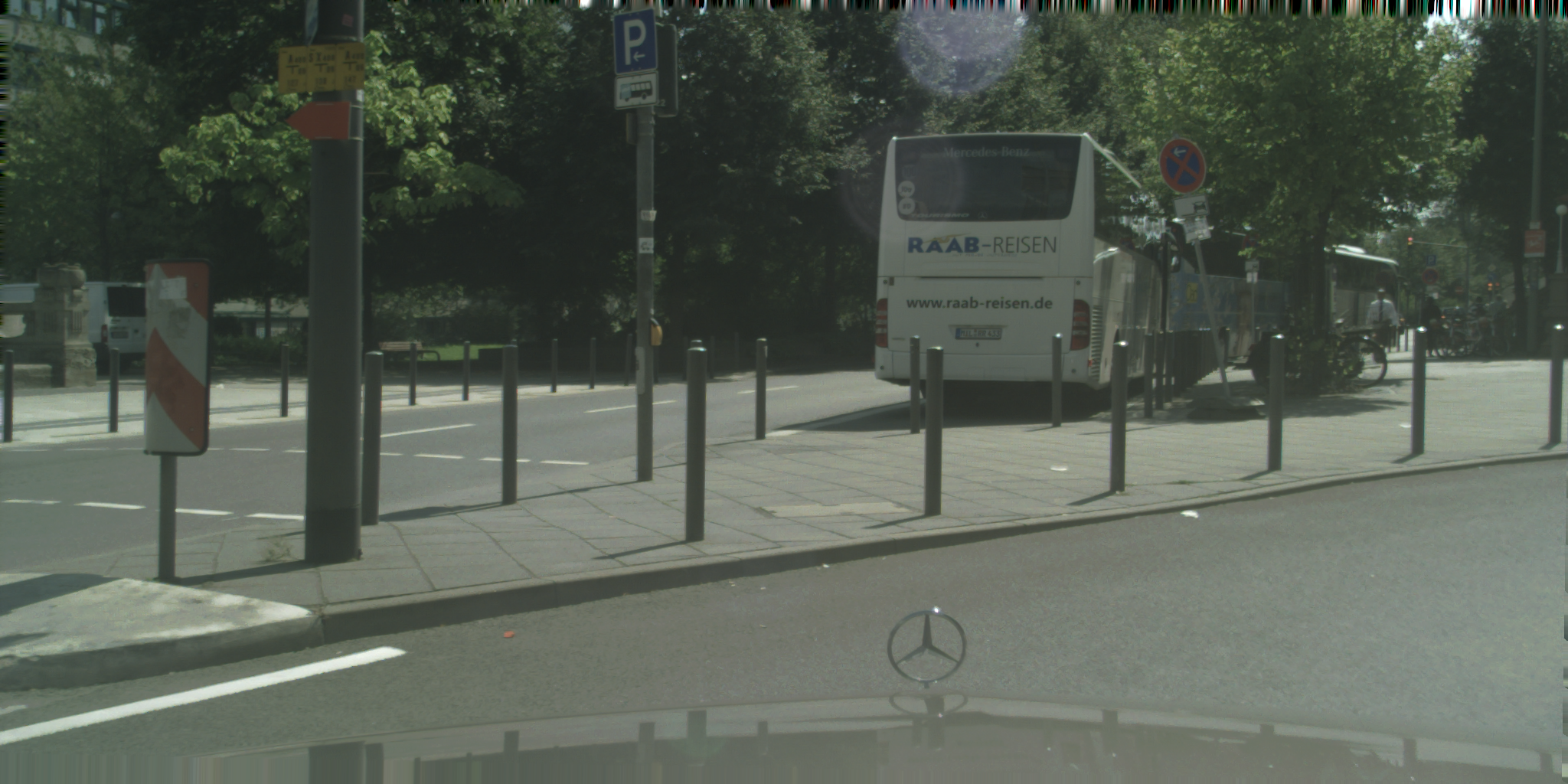}
        \caption{Image}
        \label{fig:level_img}
    \end{subfigure}
    \begin{subfigure}{0.325\textwidth}
        \centering
        \includegraphics[width=\textwidth]{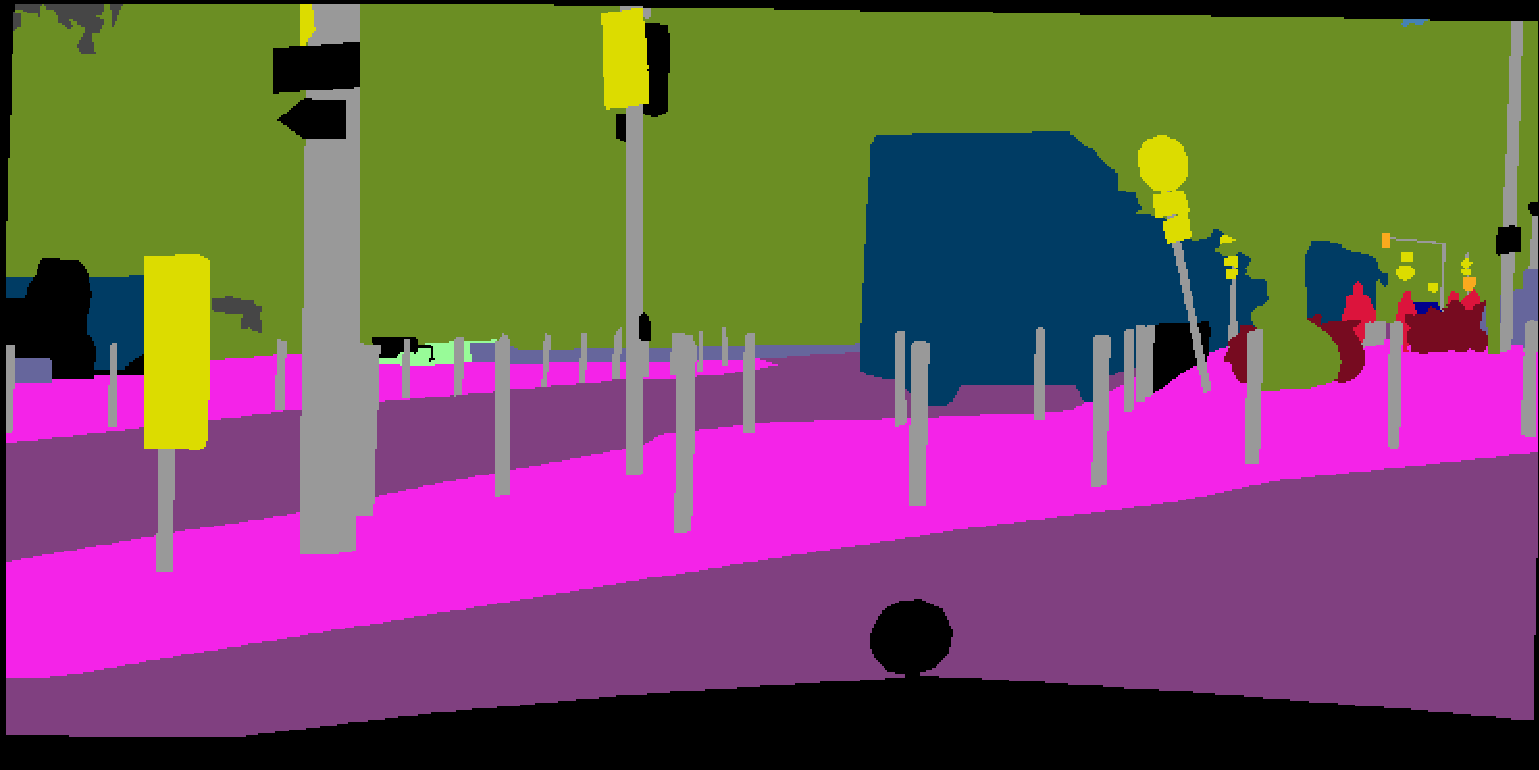}
        \caption{GT}
        \label{fig:level_target}
    \end{subfigure}
    \begin{subfigure}{0.325\textwidth}
        \centering
        \includegraphics[width=\textwidth]{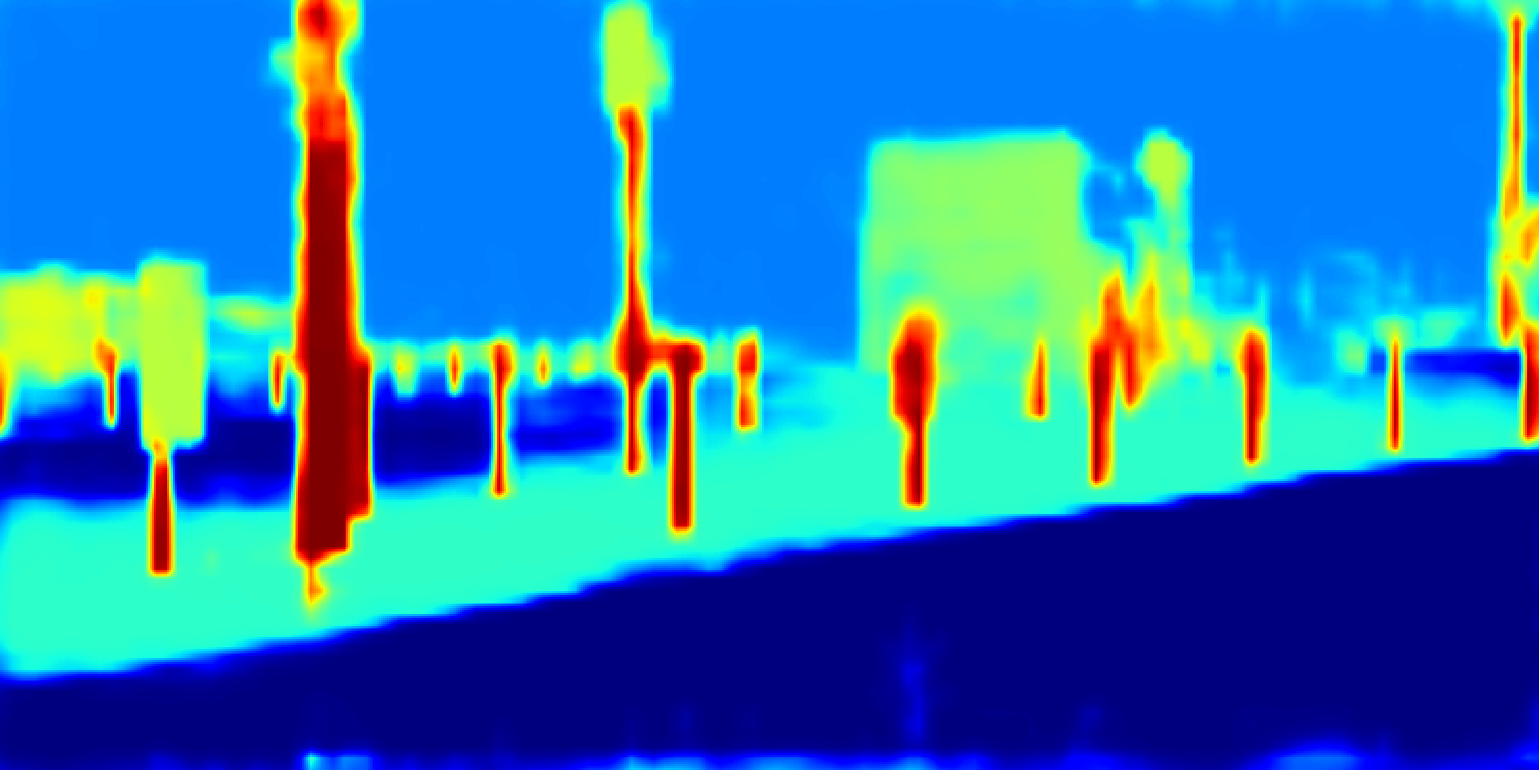}
        \caption{Semantic difficulty map}
        \label{fig:level_diff}
    \end{subfigure}
    \begin{subfigure}{0.16\textwidth}
        \centering
        \includegraphics[width=\textwidth]{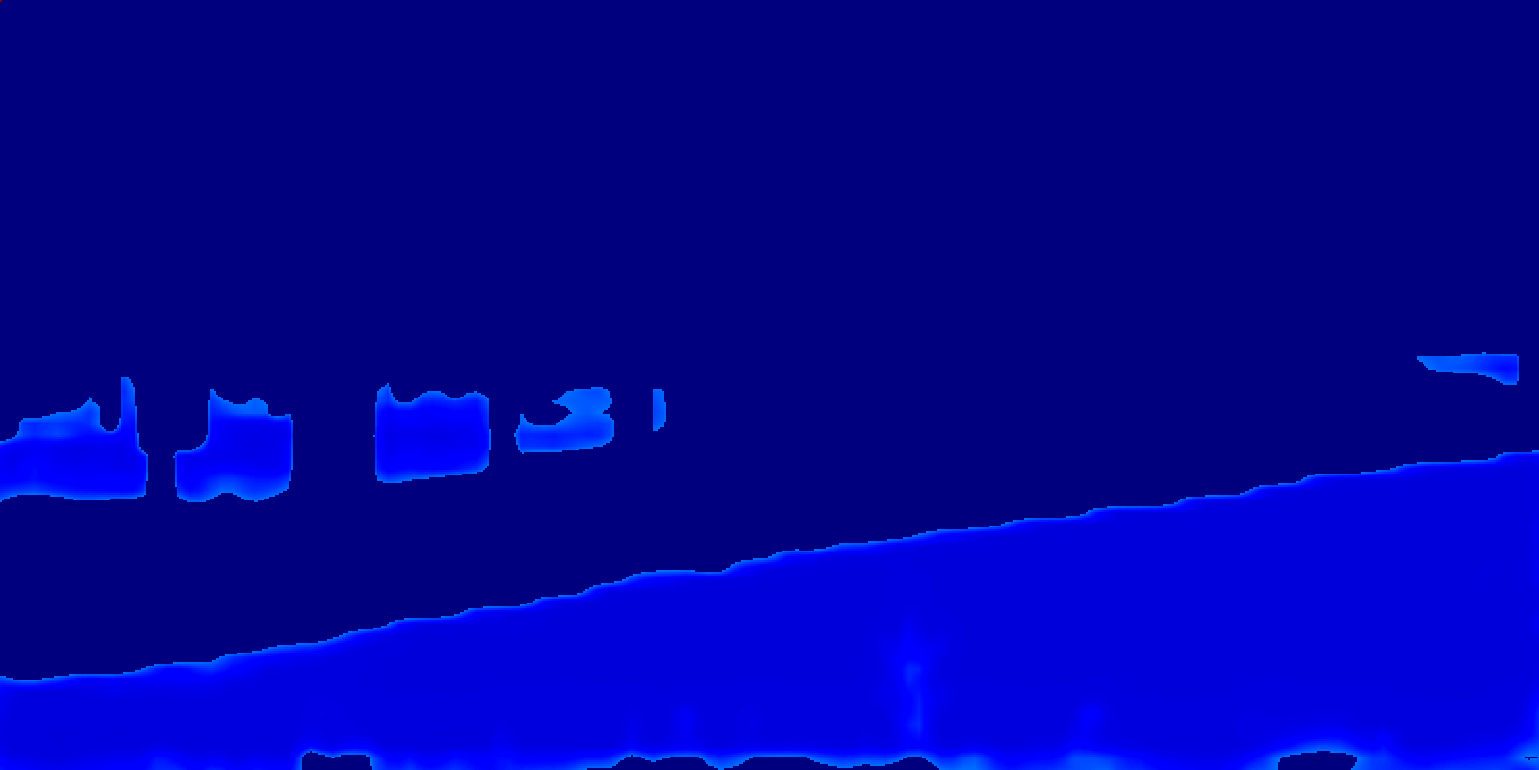}
        \caption{Level 1}
        \label{fig:diff_level1}
    \end{subfigure}
        \begin{subfigure}{0.16\textwidth}
        \centering
        \includegraphics[width=\textwidth]{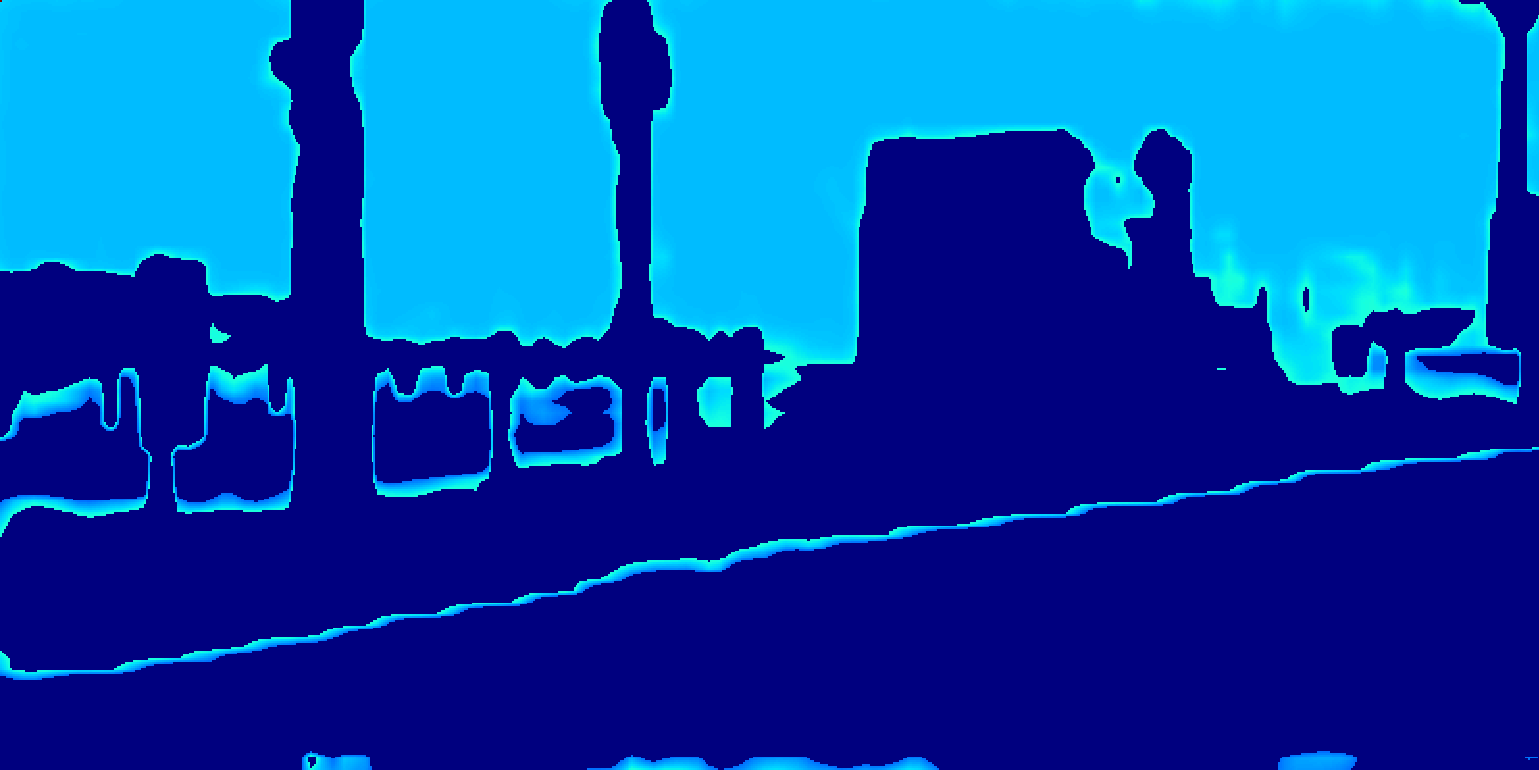}
        \caption{Level 2}
        \label{fig:diff_level2}
    \end{subfigure}
    \begin{subfigure}{0.16\textwidth}
        \centering
        \includegraphics[width=\textwidth]{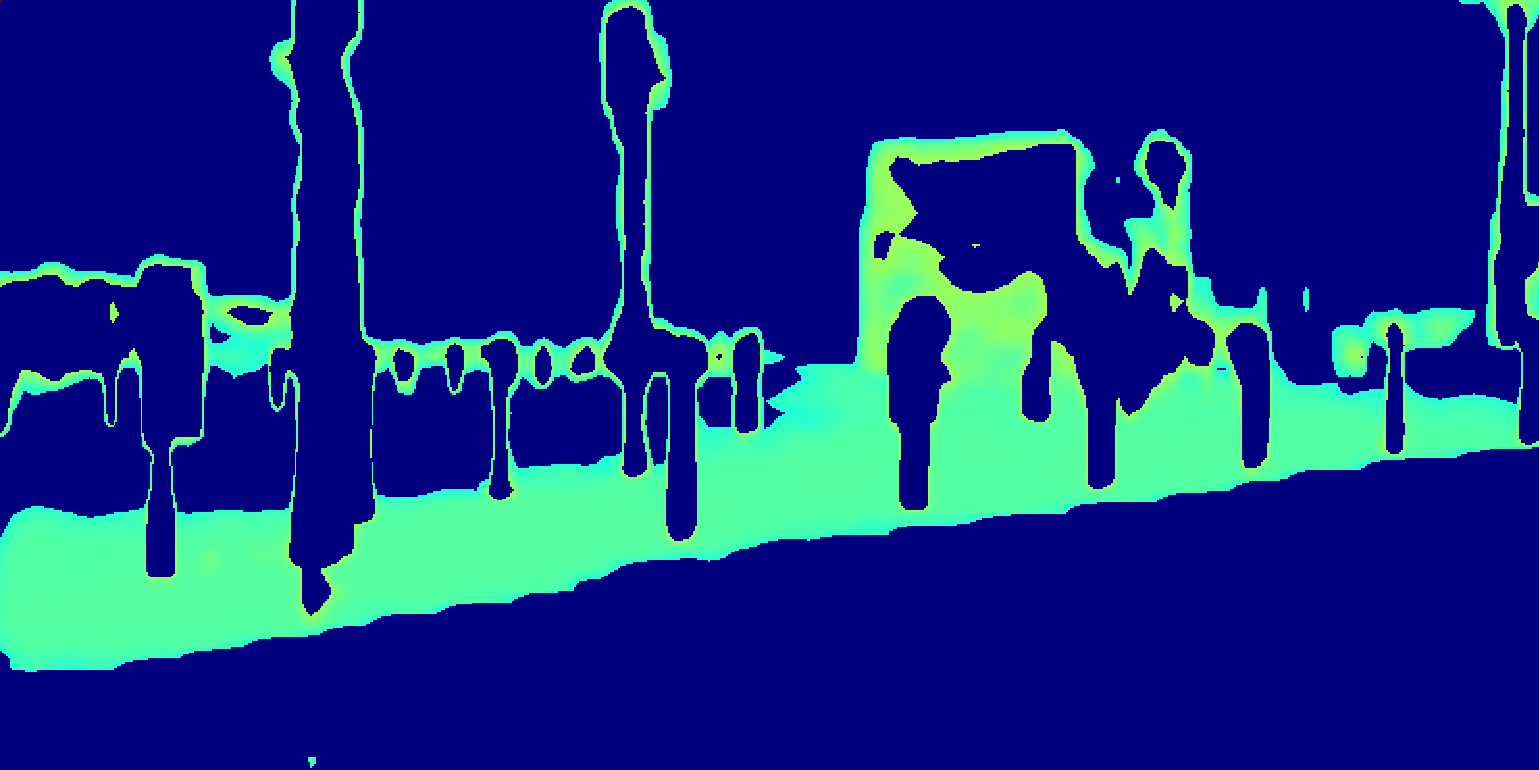}
        \caption{Level 3}
        \label{fig:diff_level3}
    \end{subfigure}
    \begin{subfigure}{0.16\textwidth}
        \centering
        \includegraphics[width=\textwidth]{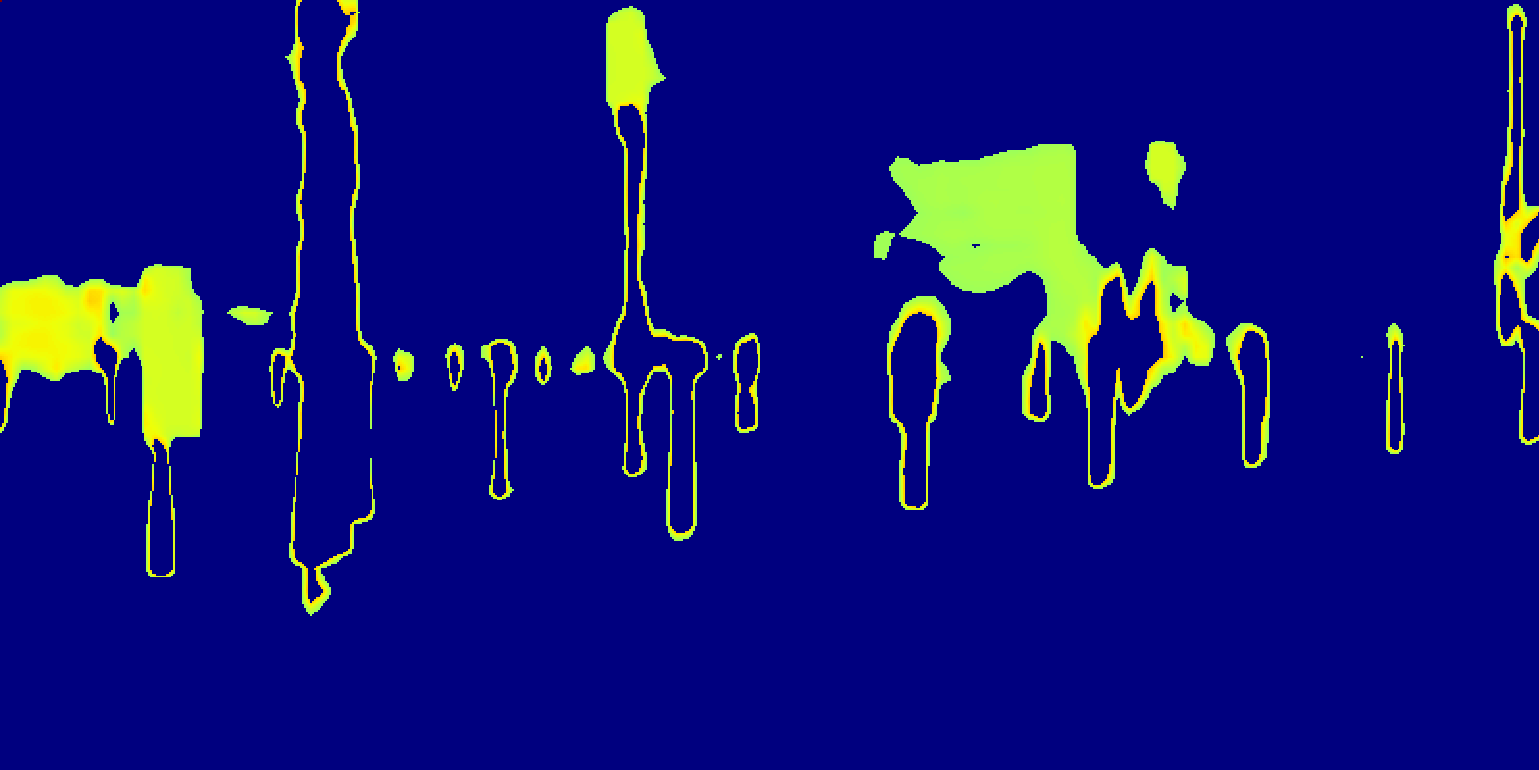}
        \caption{Level 4}
        \label{fig:diff_level4}
    \end{subfigure}
        \begin{subfigure}{0.16\textwidth}
        \centering
        \includegraphics[width=\textwidth]{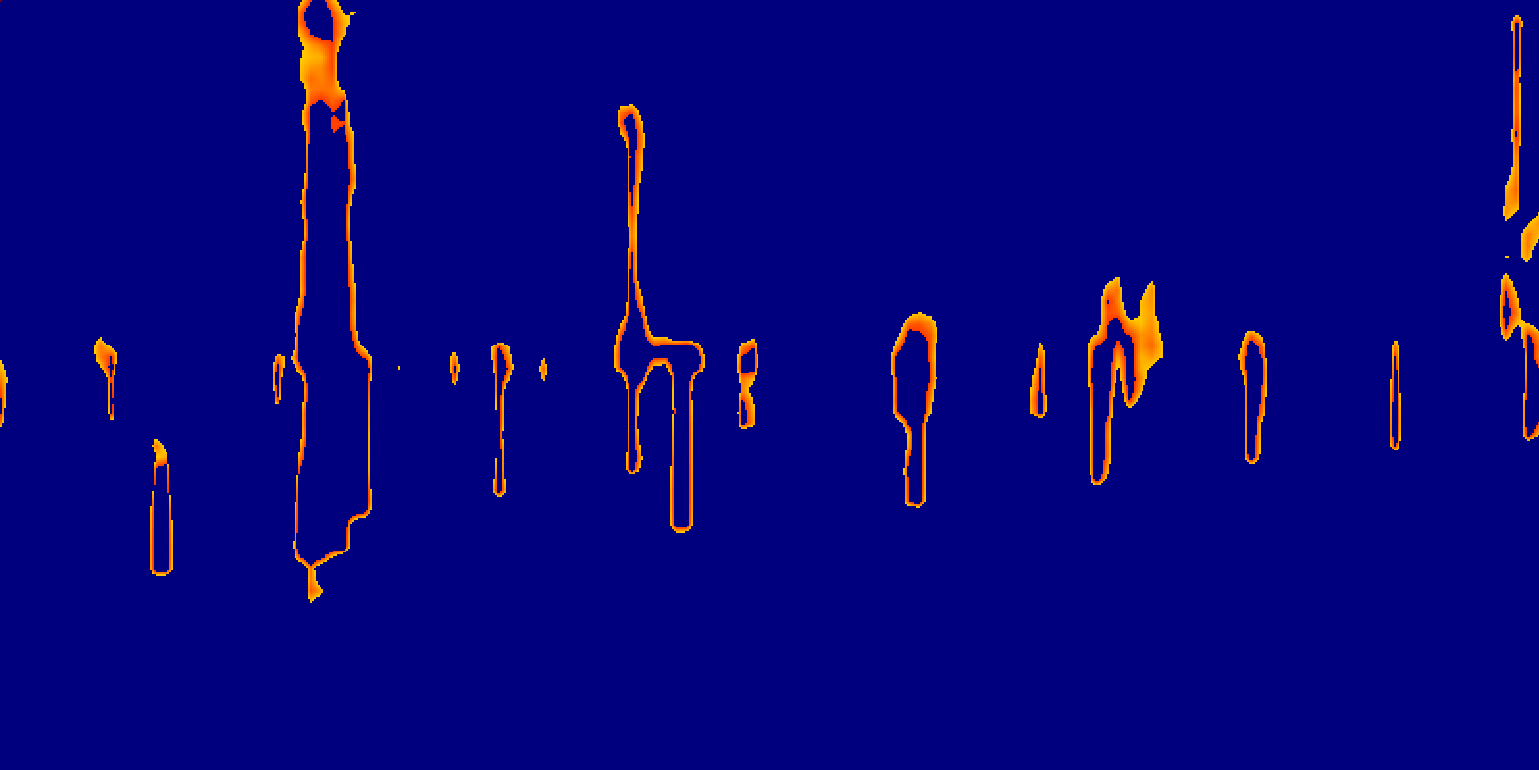}
        \caption{Level 5}
        \label{fig:diff_level5}
    \end{subfigure}
    \begin{subfigure}{0.16\textwidth}
        \centering
        \includegraphics[width=\textwidth]{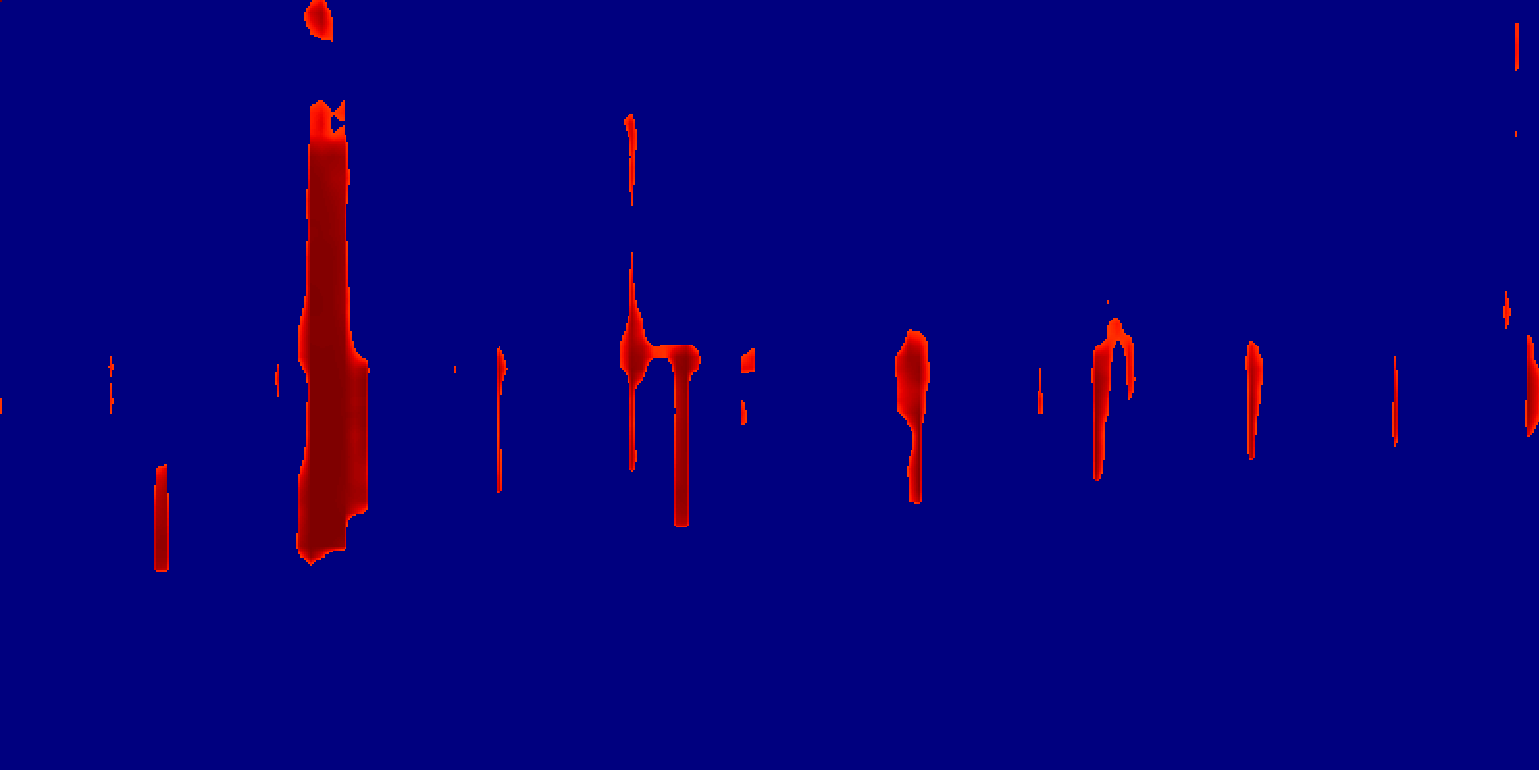}
        \caption{Level 6}
        \label{fig:diff_level6}
    \end{subfigure}
\caption{Quantified difficulty levels. Semantics in (a) are quantified into 6 difficulty levels based on (c). (d) Level 1: road. (e) Level 2: vegetation. (f) Level 3: sidewalk and a part of bus. (g) Level 4: traffic light and the rear window of bus. (h) Level 5: boundaries of poles. (i) Level 6: poles. The background of images in (d-i) is in dark blue, representing pixels not falling in the corresponding level.}
\label{fig:levels}
\end{figure}

\subsubsection{Difficulty-aware uncertainty Score (DS)}
Assume $M^c$ is the uncertainty map generated with traditional methods like Entropy, we can define the equation below to make each pixel aware of its semantic difficulty.
\begin{align} \label{equ:ds}
    \mathcal{S}_{DS} = \frac{1}{K} \sum^K_{k=1} M^c_k \cdot M_k^d,
\end{align}
where $M^c_k$ and $M^d_k$ are the uncertainty score and difficulty score of the $k^{th}$ pixel, $K$ is the total pixel number, $\mathcal{S}_{DS}$ is the average difficulty-aware uncertainty score for selecting samples with the highest uncertainty.

\subsubsection{Difficulty-aware semantic Entropy (DE)}
This acquisition function is inspired by the laddered semantic difficulty reflected on $M^d$.
Usually, pixels from the same semantic area have almost the same semantic difficulty scores, best viewed in Fig. \ref{fig:levels}(c).
In this example, we quantify the difficulty of pixels in Fig. \ref{fig:levels}(a) into 6 levels in Fig. \ref{fig:levels}(d-i), with difficulty scores gradually increasing from level 1 to level 6.
Generally, if we quantify the difficulty in an image into $L$ levels, the difficulty entropy acquisition function can be defined below to query samples with more balanced semantic difficulty, which can be viewed as a representation-based method at the image scale.
\begin{align} \label{equ:de}
    \mathcal{S}_{DE} = -\sum_{l=1}^{l=L} \frac{K_l}{K} log(\frac{K_l}{K}),
\end{align}
where $K_l$ is the number of pixels falling in the $l^{th}$ difficulty level, $K$ is the total pixel number, $L$ is the quantified difficulty levels, $\mathcal{S}_{DE}$ is the difficulty-aware semantic entropy for selecting samples with more balanced semantic difficulty. Our full algorithm of DEAL is shown in Algorithm \ref{alg:A1}.

\begin{algorithm}[t]
\renewcommand{\algorithmicrequire}{\textbf{Input:}}
\renewcommand{\algorithmicensure}{\textbf{Output:}}
    \caption{Difficulty-aware Active Learning Algorithm}
    \label{alg:A1}
    \SetAlgoLined
    \begin{algorithmic}[1] 
        \REQUIRE $\mathcal{D}^a$, $\mathcal{D}^u$, budget $m$, AL query times $N$, initialized network parameter $\Theta$
        \REQUIRE iterations $T$, weight factor $\alpha$, quantified difficulty levels $L$ (optional)
        \ENSURE $\mathcal{D}^a$, $\mathcal{D}^u$, Optimized $\Theta$
        \FOR{$n=1,2,...,N$}
            \STATE Train the two-branch difficulty learning network on $\mathcal{D}^a$
            \FOR{$t=1,2,...,T$}
            \STATE Sample $(x^a, y^a)$ from $\mathcal{D}^a$
            \STATE Compute the segmentation output $S^*$ and result $S^p$
            \STATE Obtain $M^{e}$ according to Eq. \ref{equ:mask}
            \STATE Compute the difficulty prediction $M^d$
            \STATE Compute $\mathcal{L}_{seg}$, $\mathcal{L}_{dif}$, $\mathcal{L}$ according to Eq. \ref{equ:seg}, Eq. \ref{equ:dif}, Eq. \ref{equ:sum}
            \STATE Update $\Theta$ using gradient descent
            \ENDFOR
            \STATE Rank $x^u$ based on Eq. \ref{equ:ds} or Eq. \ref{equ:de}
            \STATE Select $\mathcal{D}^s$ with top $m$ samples
            \STATE Annotate $\mathcal{D}^s$ by oralces
            \STATE $\mathcal{D}^a \leftarrow \mathcal{D}^a + \mathcal{D}^s $
            \STATE $\mathcal{D}^u \leftarrow \mathcal{D}^u - \mathcal{D}^s $
        \ENDFOR
        \RETURN $\mathcal{D}^a$, $\mathcal{D}^u$, Optimized $\Theta$
    \end{algorithmic}
\end{algorithm}


\section{Experiments and Results}

In this section, we first describe the datasets we use to evaluate our method and the implementation details, then the baseline methods, finally compare our results with these baselines.

\begin{table}[b]
    \centering
    \caption{Evaluation datasets.}
    \label{tab:dataset}
    \scalebox{0.9}{
    \begin{tabular}{ccp{1cm}<{\centering}p{1cm}<{\centering}p{1cm}<{\centering}ccc}
    \hline
    \textbf{Dataset} & \textbf{Classes} & \textbf{Train} & \textbf{Valid} & \textbf{Test} & \textbf{Initial labeled} & \textbf{Budget} & \textbf{Image Size} \\
    \hline
    CamVid \cite{brostow2009semantic} & 11 & 370 & 104 & 234 & 40 & 20 & $360\times480$ \\
    \hline
    Cityscapes \cite{cordts2016cityscapes} & 19 & 2675 & 300 & 500 & 300 & 150 & $688\times688$ \\
    \hline
    \end{tabular}
    }
\end{table}

\subsection{Experimental Setup}

\subsubsection{Datasets} 
We evaluate DEAL on two street scene semantic segmentation datasets: CamVid \cite{brostow2009semantic} and Cityscapes \cite{cordts2016cityscapes}.
For Cityscapes, we randomly select 300 samples from the training set as the validation set, and the original validation set serves as the test set, same to \cite{sinha2019variational}.
The detailed configurations are list in Table \ref{tab:dataset}.
For each dataset, we first randomly sample $10\%$ data from the training set as the initial annotated dataset $\mathcal{D}^a$, then iteratively query $5\%$ new data $\mathcal{D}^s$ from the remaining training set, which serves as the unlabeled data pool $\mathcal{D}^u$. Considering samples in the street scenes have high similarities, we first randomly choose a subset from $\mathcal{D}^u$, then query $m$ samples from the subset, same to \cite{beluch2018power}.

\subsubsection{Implementation Details}
We adopt the Deeplabv3+ \cite{chen2018encoder} architecture with a Mobilenetv2 \cite{sandler2018mobilenetv2} backbone. 
For each dataset, we use mini-batch SGD \cite{krizhevsky2012imagenet} with momentum 0.9 and weight decay $5e^{-4}$ in training. The batch size is 4 and 8 for CamVid and Cityscapes, respectively.
For all methods and the upper bound method with the full training data, we train 100 epochs with an unweighted cross-entropy loss function.
Similar to \cite{chen2018encoder}, we apply the ``poly" learning rate strategy and the initial learning rate is 0.01 and multiplied by $(1-\frac{iter}{max\_iter})^{0.9}$.
To accelerate the calculation of the probability attention module, the input of the difficulty branch is resized to $80\times60$ and $86\times86$ for CamVid and Cityscapes.

\subsection{Evaluated Methods}

We compare DEAL with the following methods.
\emph{Random} is a simple baseline method.
\emph{Entropy} and \emph{QBC} are two uncertainty-based methods.
\emph{Core-set} and \emph{VAAL} are two representation-based methods.
\emph{DEAL (DS)} and \emph{DEAL (DE)} are our methods with different acquisition functions.

\begin{itemize}
    \item \textbf{Random}: each sample in $\mathcal{D}^u$ is queried with uniform probability.
    \item \textbf{Entropy} (Uncertainty): we query samples with max mean entropy of all pixels. \cite{yoo2019learning} and \cite{Casanova2020Reinforced} have verified this method is quite competitive in image classification and segmentation tasks.
    \item \textbf{QBC} (Uncertainty): previous methods designed for semantic segmentation, like \cite{yang2017suggestive,gorriz2017cost,tan2019batch,mackowiak2018cereals}, all use a group of models to measure uncertainty. We use the efficient MC dropout to represent these methods and report the best performance out of both the max-entropy and variation-ratio acquisition functions.
    \item \textbf{Core-set} (Representation): we query samples that can best cover the entire data distribution. We use the global average pooling operation on the encoder output features of Deeplabv3+ and get a feature vector for each sample. Then \textit{k-Center-Greedy} strategy is used to query the most informative samples, and the distance metric is $l_2$ distance according to \cite{sener2017active}.
    \item \textbf{VAAL} (Representation): as a new state-of-the-art task-agnostic method, the sample query process of VAAL is totally separated from the task learner. We use this method to query samples that are most likely from $\mathcal{D}^u$ and then report the performance with our task model.
    \item \textbf{DEAL (DS)}: out method with DS acquisition function. We employ \emph{Entropy} uncertainty maps in our experiments.
    \item \textbf{DEAL (DE)}: out method with DE acquisition function. The quantified difficulty levels $L$ are set 8 and 10 for CamVid and Cityscapes, respectively. 
\end{itemize}

\subsection{Experimental Results}

The mean Intersection over Union (mIoU) at each AL stage: 10\%, 15\%, 20\%, 25\%, 30\%, 35\%, 40\% of the full training set, are adopted as the evaluation metric. Every method is run 5 times and the average mIoUs are reported.

\begin{figure}[t]
    \centering
    \begin{subfigure}[b]{0.49\columnwidth}
        \centering
        \includegraphics[width=\columnwidth]{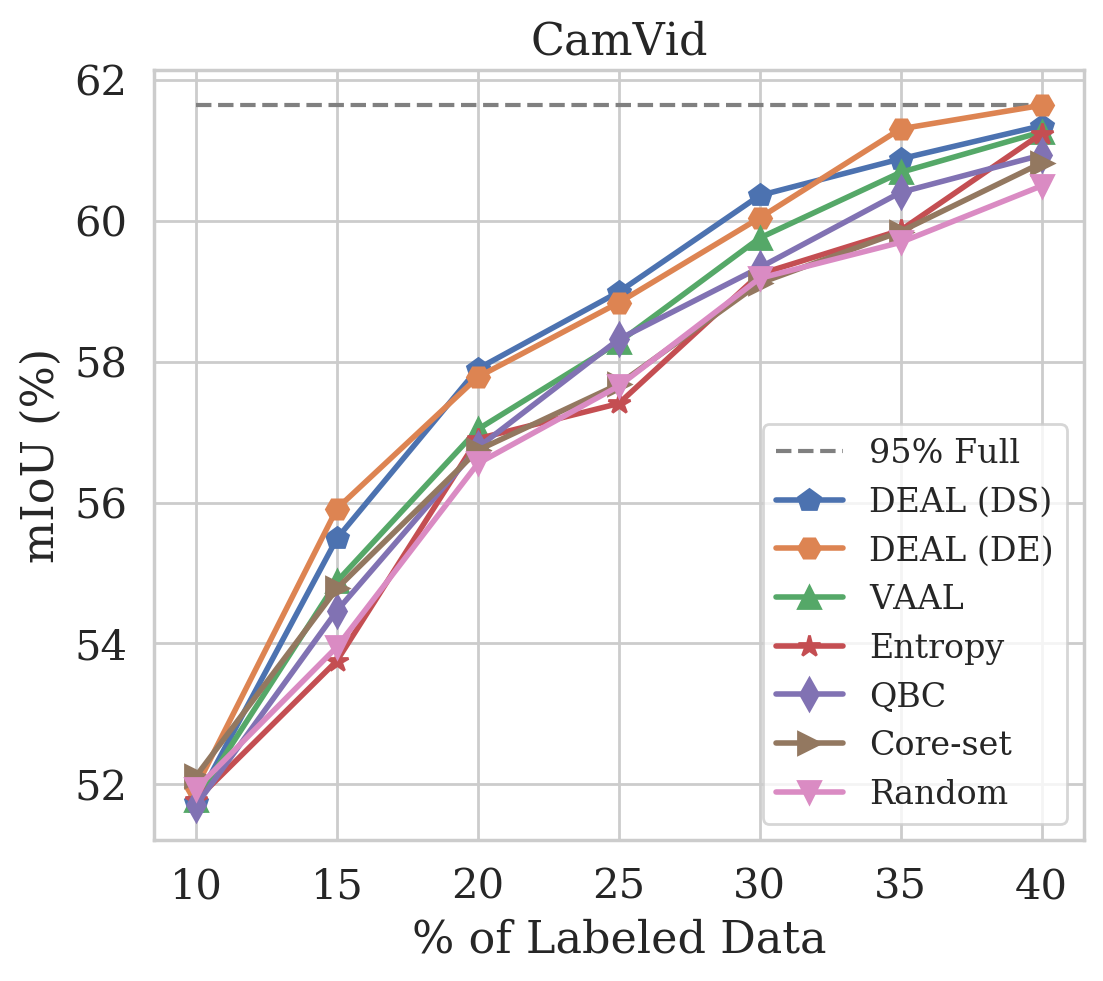}
        \caption{AL in CamVid}
        \label{fig:res_cam}
    \end{subfigure}
    \begin{subfigure}[b]{0.49\columnwidth}
        \centering
        \includegraphics[width=\columnwidth]{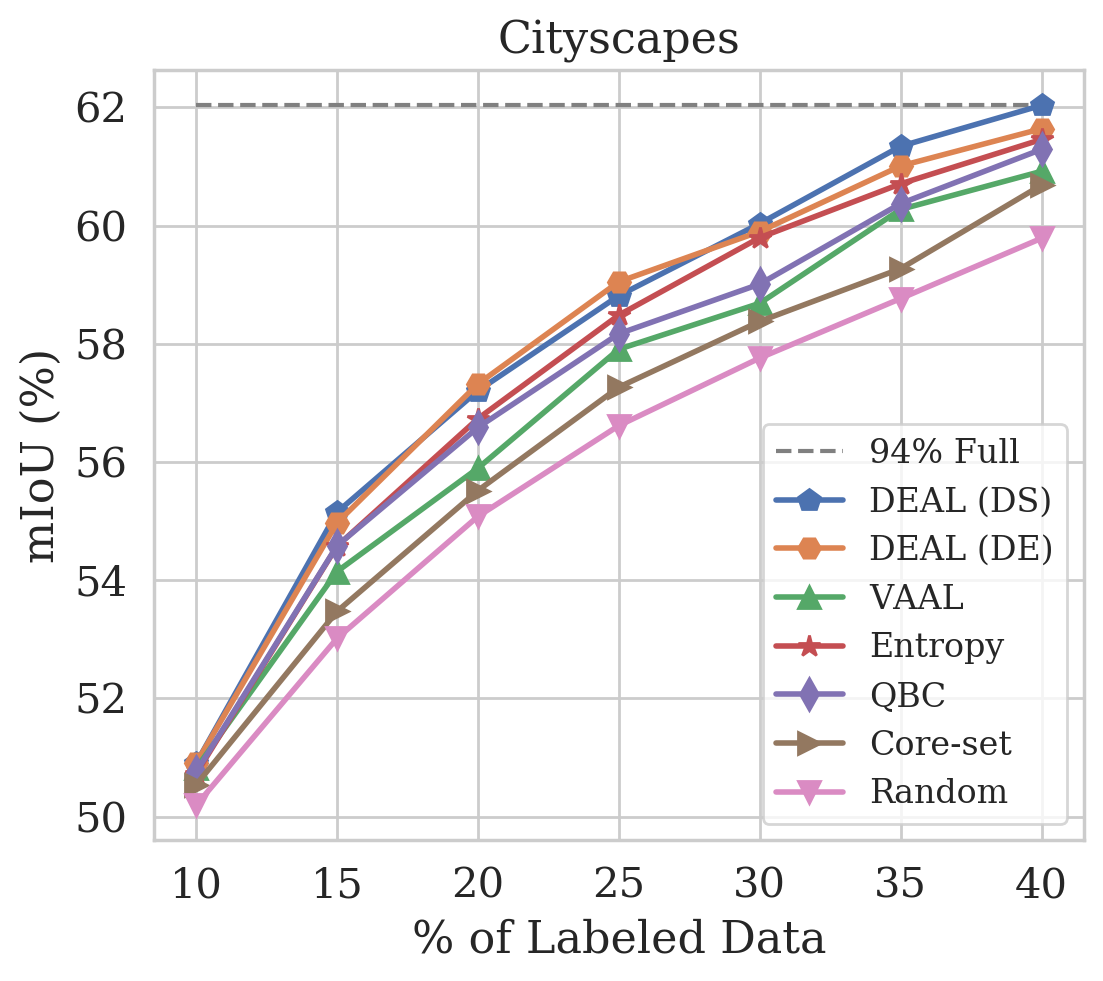}
        \caption{AL in Cityscapes}
        \label{fig:res_cit}
    \end{subfigure}
    \caption{
    DEAL performance on CamVid \cite{brostow2009semantic} and Cityscapes \cite{cordts2016cityscapes}.
    Every method is evaluated by the average mIoU of 5 runs. The dashed line represents the upper performance we can reach compared with the full training data.
    }
\label{fig:res}
\end{figure}

\subsubsection{Results on CamVid}
Fig. \ref{fig:res}(a) shows results on a small dataset CamVid.
Both \emph{DEAL (DS)} and \emph{DEAL (DE)} outperform baseline methods at each AL stage. We can obtain a performance of 61.64\% mIoU with 40\% training data, about 95\% of the upper performance with full training data.
\emph{Entropy} can achieve good results at the last stage. However, it's quite unstable and depends much on the performance of current model, making it behave poorly and exceeded by \emph{Random} at some stages. 
On the contrary, \emph{DEAL (DS)} behaves better with the difficulty attention.
\emph{QBC} has a more stable growth curve as it depends less on the single model.
Representation-based methods like \emph{VAAL} and \emph{Core-set} behave much better at earlier stages like 15\% and 20\%. However, \emph{Core-set} lags behind later while \emph{VAAL} still works well. 
Also, the experiment results suggest that the data diversity is more important when the entire dataset is small.

\subsubsection{Results on Cityscapes}
Fig. \ref{fig:res}(b) shows results on a larger dataset Cityscapes.
The budget is 150 and all methods have more stable growth curves. 
When the budget is enough, \emph{Entropy} can achieve better performance than other baseline methods.
Consistently, with semantic difficulty, both \emph{DEAL (DS)} and \emph{DEAL (DE)} outperform other methods. Table \ref{tab:cit} shows the per-class IoU for each method at the last AL stage (40\% training data). Compared with \emph{Entropy}, our method are more competitive on the difficult classes, such as \emph{pole}, \emph{traffic sign}, \emph{rider} and \emph{motorcycle}.
For representation-based methods, the gap between \emph{Core-set} and \emph{VAAL} is more obvious, suggesting that \emph{Core-set} is less effective when the input has a higher dimension. And \emph{VAAL} is easily affected by the performance of the learned variational autoencoder, which introduces more uncertainty to the active learning system.
If continue querying new data, our method will reach the upper performance of full training data with about 60\% data.

\begin{table}[t]
    \centering
    \caption{Per-class IoU of and mIoU [\%] on Cityscapes original validation set with 40\% training data. For clarity, only the average of 5 runs are reported, and the best and the second best results are highlighted in \textbf{bold} and \underline{\textbf{underline bold}}.}
    \label{tab:cit}
    \scalebox{0.8}{
    \begin{tabular}{ >{\centering\arraybackslash}m{2cm}
                 *{10}{>{\centering\arraybackslash}m{1.2cm}}}
    \toprule
    \textbf{Method} & \textbf{Road} & \textbf{Side-walk} & \textbf{Build-ing} & \textbf{Wall} & \textbf{Fence} & \textbf{Pole} & \textbf{Traffic Light} & \textbf{Traffic Sign} & \textbf{Vege-tation} & \textbf{Terrain} \\
    \midrule
    Random    & 96.03 & 72.36 & 86.79 & 43.56 & 44.22 & 36.99 & 35.28 & 53.87 & 86.91 & 54.58  \\ 
    Core-set  & 96.12 & 72.76 & 87.03 & 44.86 & \textbf{45.86} & 35.84 & 34.81 & 53.07 & 87.18 & 53.49 \\
    VAAL      & \underline{\textbf{96.22}} & \underline{\textbf{73.27}} & 86.95 & \textbf{47.27} & 43.92 & 37.40 & 36.88 & 54.90 & 87.10 & 54.48 \\
    QBC       & 96.07 & 72.27 & 87.05 & \underline{\textbf{46.89}} & 44.89 & 37.21 & 37.57 & 54.53 & 87.51 & 55.13 \\
    Entropy   & \textbf{96.28} & \textbf{73.31} & \textbf{87.13} & 43.82 & 43.87 & 38.10 & \textbf{37.74} & 55.39 & \underline{\textbf{87.52}} & 53.68 \\
    DEAL (DS) & 96.21 & 72.72 & 86.94 & 46.11 & 44.22 & \underline{\textbf{38.18}} & \underline{\textbf{37.62}} & \textbf{55.66} & 87.34 & \underline{\textbf{55.62}} \\
    DEAL (DE) & 95.89 & 71.69 & \underline{\textbf{87.09}} & 45.61 & \underline{\textbf{44.94}} & \textbf{38.29} & 36.51 & \underline{\textbf{55.47}} & \textbf{87.53} & \textbf{56.90} \\
    \bottomrule
    \textbf{} & \textbf{Sky} & \textbf{Pedes-trian} & \textbf{Rider} & \textbf{Car} & \textbf{Truck} & \textbf{Bus} & \textbf{Train} & \textbf{Motor-cycle} & \textbf{Bicycle} & \textbf{mIoU} \\
    \midrule
    Random    & 91.47 & 62.74 & 37.51 & 88.05 & 54.64 & 61.00 & 43.69 & 30.58 & 55.67 & 59.79 \\
    Core-set  & 91.89 & 62.48 & 36.28 & 87.63 & 57.25 & \underline{\textbf{67.02}} & \textbf{56.59} & 29.34 & 53.56 & 60.69 \\
    VAAL      & 91.63 & 63.44 & 38.92 & 87.92 & 50.15 & 63.70 & 52.36 & 35.99 & 54.97 & 60.92 \\
    QBC       & 91.87 & 63.79 & 38.76 & 88.04 & 53.88 & 65.92 & \underline{\textbf{54.32}} & 32.68 & 56.15 & 61.29 \\
    Entropy   & \underline{\textbf{92.05}} & \underline{\textbf{63.96}} & 38.44 & \textbf{88.38} & \underline{\textbf{59.38}} & 64.64 & 50.80 & 36.13 & \textbf{57.10} & 61.46 \\
    DEAL (DS) & \textbf{92.10} & 63.92 & \textbf{40.39} & 87.87 & \textbf{59.85} & \textbf{67.32} & 52.30 & \underline{\textbf{38.88}} & 55.44 & \textbf{62.04} \\
    DEAL (DE) & 91.78 & \textbf{64.25} & \underline{\textbf{39.77}} & \underline{\textbf{88.11}} & 56.87 & 64.46 & 50.39 & \textbf{38.92} & \underline{\textbf{56.69}} &  \underline{\textbf{61.64}} \\
    \bottomrule
    \end{tabular}
    }
\end{table}


\section{Ablation Study}

\subsection{Effect of PAM}
\label{sec:pam}

\begin{figure}[b]
    \centering
    \begin{overpic}[width=1.\linewidth]{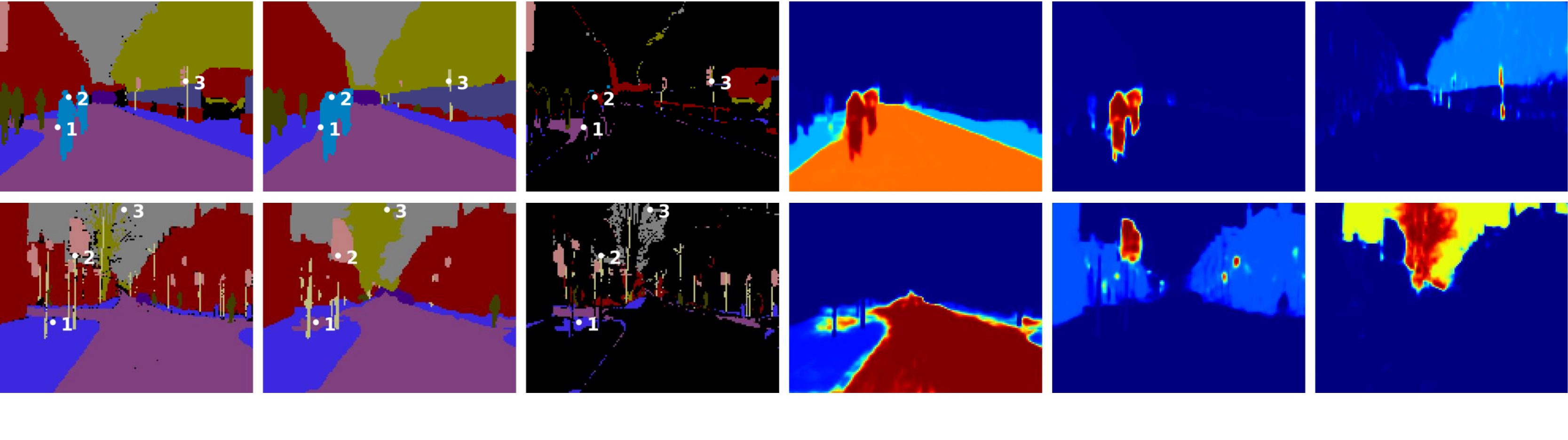}
    \put(4,0){\fontsize{8}{8} \selectfont (a) GT}
    \put(16,0){\fontsize{8}{8} \selectfont (b) Prediction}
    \put(32.5,0){\fontsize{8}{8} \selectfont (c) Error mask}
    \put(49,0){\fontsize{8}{8} \selectfont (d) Attention 1}
    \put(66,0){\fontsize{8}{8} \selectfont (e) Attention 2}
    \put(83,0){\fontsize{8}{8} \selectfont (f) Attention 3}
    \end{overpic}
    \caption{Attention heatmaps of three selected wrong predicted pixels on CamVid \cite{brostow2009semantic}. (a) GT. (b) Prediction. (c) Error mask. (d,e,f) Attention heatmaps of the three selected points, which are marked as \{1,2,3\} in (a,b,c). The warmer color means the more dependency.
    }
    \label{fig:att}
\end{figure}
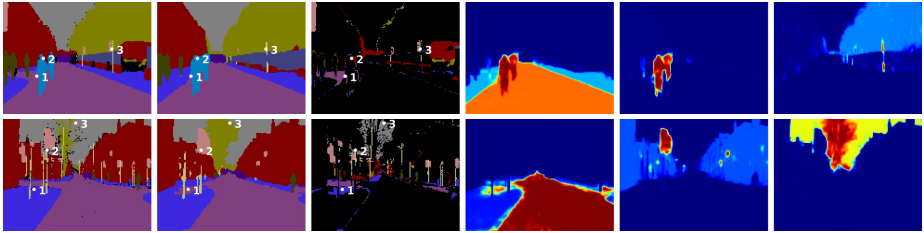

To further understand the effect of PAM, we first visualize the attention heatmaps of the wrong predictions in Fig. \ref{fig:att}. For each row, three points are selected from \emph{error mask} and marked as $\{1,2,3\}$ in Fig. \ref{fig:att}(a,b,c). 
In the first row, point 1 is from \emph{road} and misclassified as \emph{bicyclist}, we can observe that its related classes are \emph{bicyclist}, \emph{road} and \emph{sidewalk} in Fig. \ref{fig:att}(d).
Point 2 is from \emph{buildings} and misclassified as \emph{bicyclist}, too. 
Point 3 is from \emph{sign symbol} and misclassified as \emph{tree}, we can also observe its related semantic areas in Fig. \ref{fig:att}(f). 

\begin{figure}[t]
    \centering
    \begin{subfigure}[b]{0.48\columnwidth}
        \centering
        \includegraphics[width=\columnwidth]{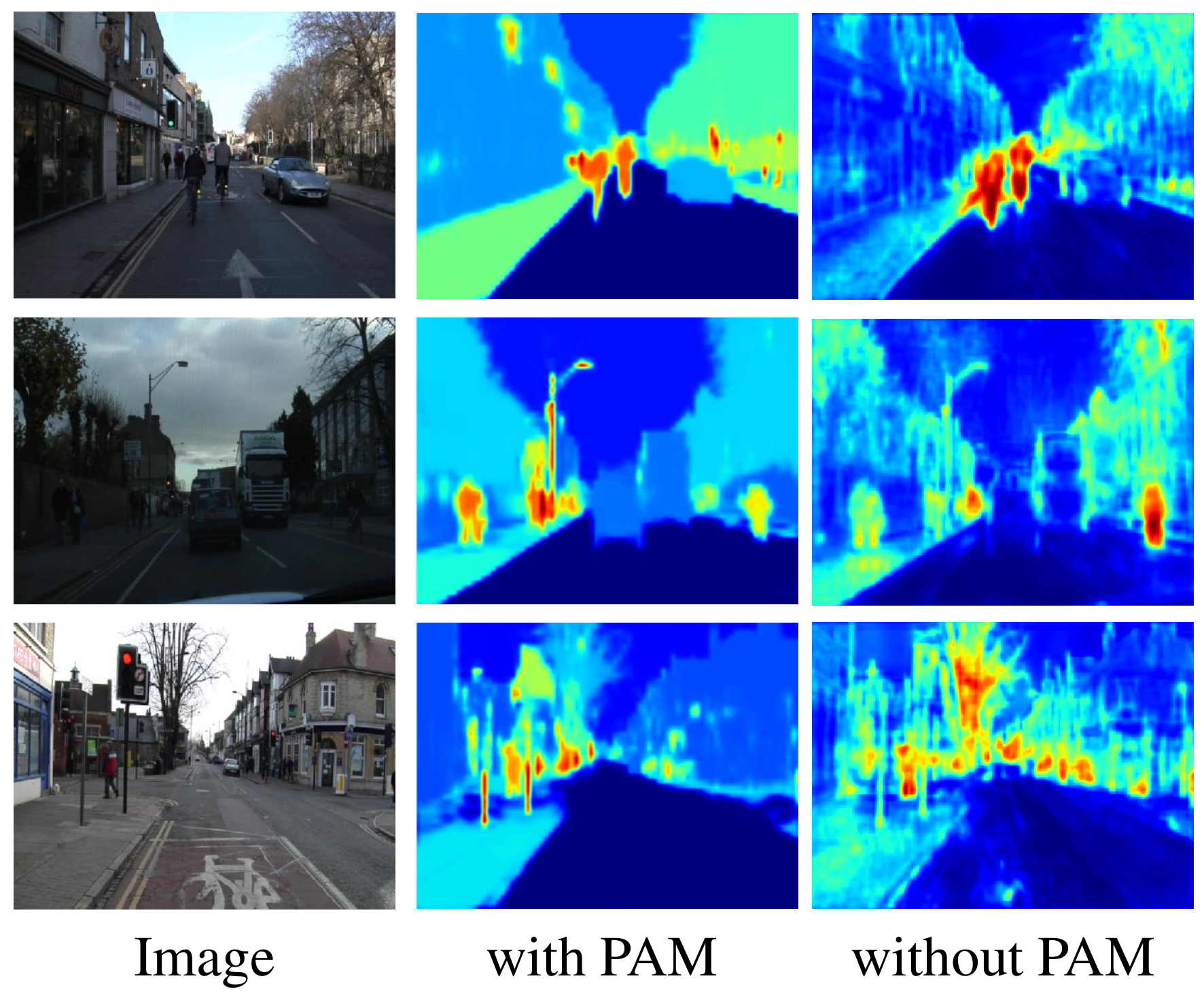}
        \caption{Semantic difficulty maps with/o PAM}
        \label{fig:uncertain_pam}
    \end{subfigure}
    \begin{subfigure}[b]{0.48\columnwidth}
        \centering
        \includegraphics[width=\columnwidth]{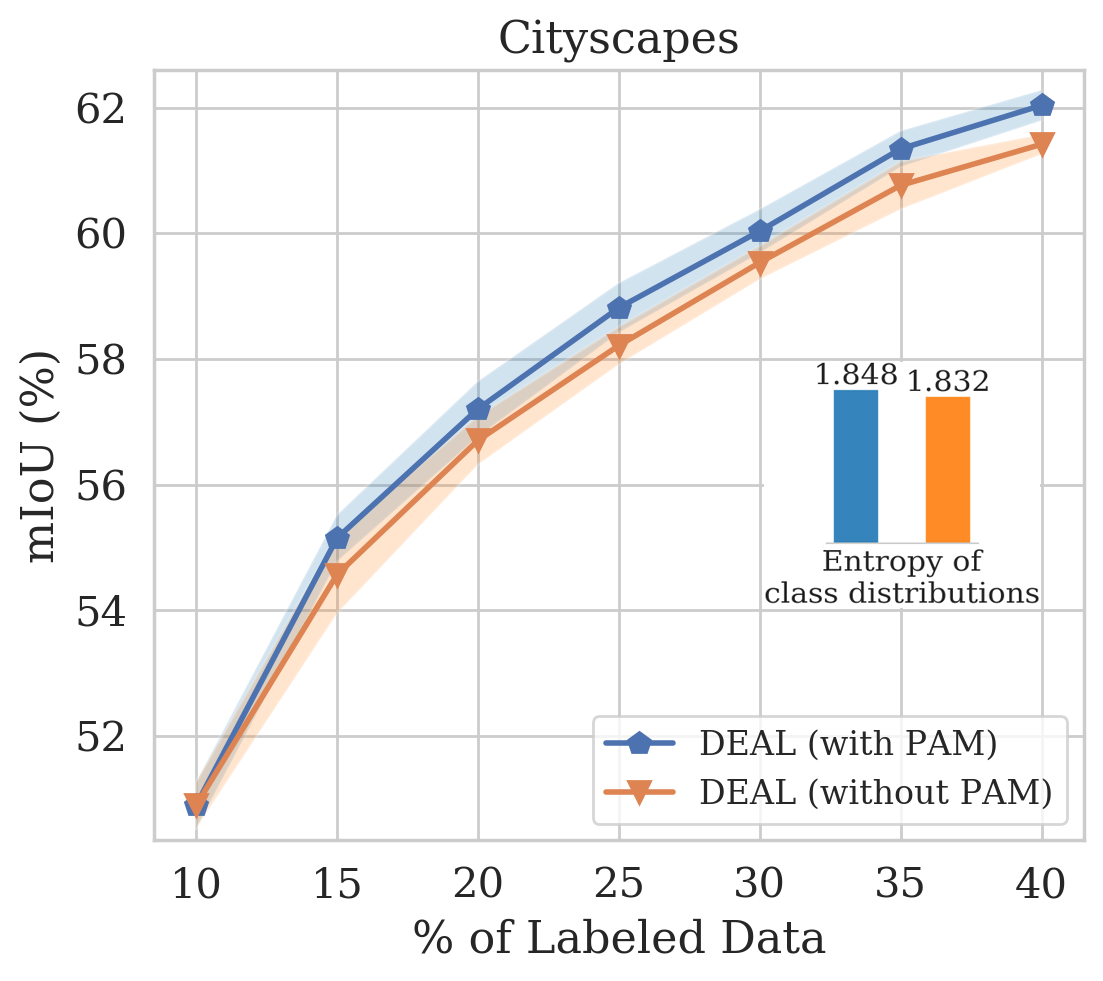}
        \caption{AL in Cityscapes with/o PAM}
        \label{fig:res_pam}
    \end{subfigure}
    \caption{Ablation study on PAM. (a) The first column are images from CamVid \cite{brostow2009semantic}, the second and third columns are semantic difficulty maps learned with and without PAM. (b) DEAL performance on Cityscapes \cite{cordts2016cityscapes} with and without PAM.
    We report the mean and standard deviation of 5 runs and the average entropy of class distributions of all AL stages.
    }
\label{fig:pam}
\end{figure}

Then we conduct an ablation study by removing PAM and directly learning the semantic difficulty map without the attention among pixels.
The qualitative results are shown in Fig. \ref{fig:pam}(a).
Basically, without the long-range dependencies, pixels of the same semantic can learn quite different scores because the learned score of each pixel is more sensitive to its original uncertainty.
Combined with PAM, we can learn more smooth difficulty map, which is more friendly to the annotators since the aggregated semantic areas are close to the labeling units in the real scenario.
Also, we compare this ablation model with our original model on Cityscapes in Fig. \ref{fig:pam} (b). DEAL with PAM can achieve a better performance at each AL stage. DEAL without PAM fails to find samples with more balanced semantic difficulty, which makes it get a lower entropy of class distributions.

\subsection{Branch Position}
\label{sec:branch_pos}
\begin{figure}[t]
    \centering
    \begin{subfigure}[b]{0.42\textwidth}
        \centering
        \includegraphics[width=\columnwidth]{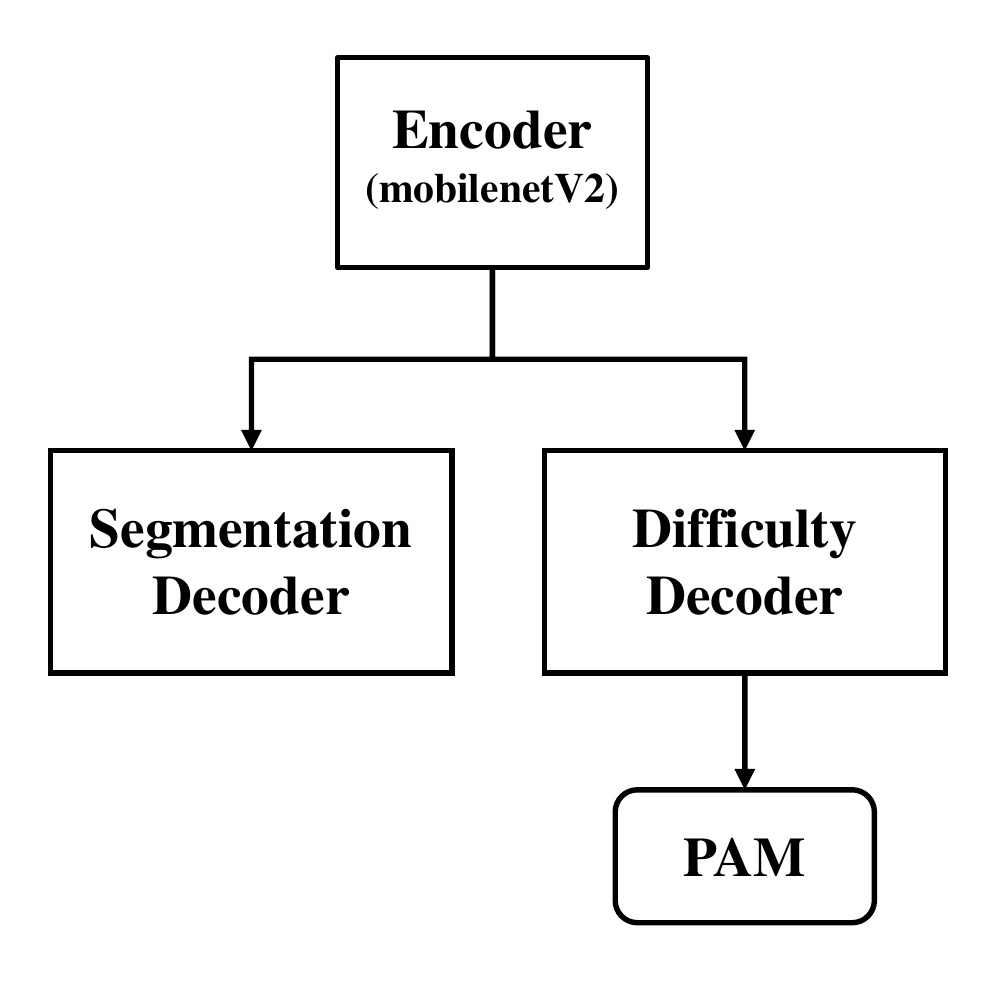}
        \caption{Modified model architecture}
    \end{subfigure}
    \begin{subfigure}[b]{0.48\textwidth}
        \centering
        \includegraphics[width=\columnwidth]{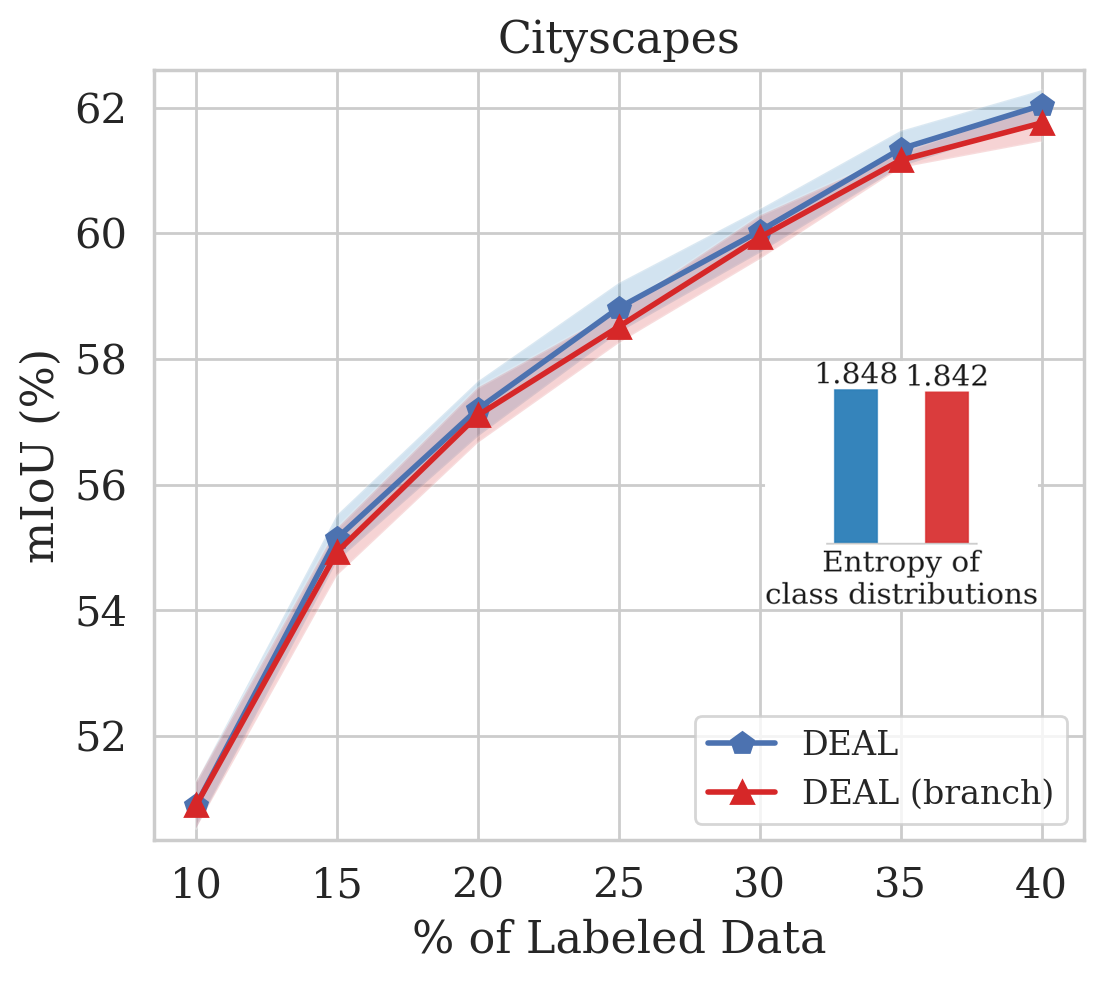}
        \caption{AL in Cityscapes}
    \end{subfigure}
    \caption{Ablation study on branch position. (a) Modified model architecture. (b) DEAL performance on Cityscapes with different model architectures. \emph{DEAL (branch)} is the modified version. We also report the mean and standard deviation of 5 runs and the average entropy of class distributions of all AL stages.
    }
    \label{fig:branch}
\end{figure}

In this section, we discuss the branch position of our framework.
In our method above, the semantic difficulty branch is simply added after the segmentation branch. It may occur to us that if the segmentation branch performs poorly, the difficulty branch will perform poorly, too. These two tasks should be separated earlier and learn independent features. Thus, we modify our model architecture and branch out two tasks earlier at the boarder of encoder and decoder based on the Deeplabv3+\cite{chen2018encoder} architecture, as shown in Fig. \ref{fig:branch}(a). 
Also, we compare the AL performance on Cityscapes with both architectures in Fig. \ref{fig:branch}(b). The performance of the modified version is slightly poor than the original version but still competitive with other methods.
However, this modified version requires more computations, while our original version is simple yet effective and can be easily plugged into any segmentation networks.

\section{Conclusion and Future Work}

In this work, we have introduced a novel Difficulty-awarE Active Learning (DEAL) method for semantic segmentation, which incorporates the semantic difficulty to select the most informative samples.
For any segmentation network, the \textit{error mask} is firstly generated with the predicted segmentation result and GT. Then, with the guidance of \textit{error mask}, the probability attention module is introduced to aggregate similar pixels and predict the semantic difficulty maps.
Finally, two acquisition functions are devised. One is combining the uncertainty of segmentation result and the semantic difficulty. The other is solely based on the difficulty.
Experiments on CamVid and Cityscapes demonstrate that the proposed DEAL achieves SOTA performance and can effectively improve the performance of hard semantic areas. 
In the future work, we will explore more possibilities with the semantic difficulty map and apply it to region-level active learning method for semantic segmentation.


\section{Acknowledgments}

This work is funded by National Key Research and Development Project (Grant No: 2018AAA0101503) and State Grid Corporation of China Scientific and Technology Project: Fundamental Theory of Human-in-the-loop Hybrid-Augmented Intelligence for Power Grid Dispatch and Control.

\bibliographystyle{splncs}
\bibliography{egbib}

\end{document}